\documentclass[11pt,a4paper]{article}
\usepackage[hyperref]{eacl2021}
\usepackage{times}
\usepackage{latexsym}

\usepackage{url}
\usepackage[T1]{fontenc}
\usepackage{listings}
\usepackage{rotating}
\usepackage{tablefootnote,footnotehyper}
\usepackage{algorithm2e}
\usepackage{color, colortbl}
\usepackage{amssymb}
\usepackage{booktabs} 
\usepackage{enumitem}
\usepackage{multirow}
\usepackage{graphicx}
\usepackage{subfigure}
\usepackage{amsmath}
\usepackage{floatrow}
\usepackage{pdflscape}
\usepackage{pifont}
\usepackage{supertabular}
\usepackage{siunitx}
\newcommand{\cmark}{\ding{51}}%
\newcommand{\xmark}{\ding{55}}%
\newcommand{\Ni}{({\em i})~}
\newcommand{\Nii}{({\em ii})~}
\newcommand{\Niii}{({\em iii})~}
\newcommand{\Niv}{({\em iv})~}
\newcommand{\Nv}{({\em v})~}

\definecolor{cadmiumgreen}{rgb}{0.0, 0.42, 0.24}
\definecolor{carnelian}{rgb}{0.7, 0.11, 0.11}
\definecolor{gold}{rgb}{0.72, 0.53, 0.04}

\newcommand{\flores}{\textsc{Flores-101}}

\newif\ifshowadds
\showaddstrue 

\usepackage{microtype}

\aclfinalcopy 

\title{The \flores{} Evaluation Benchmark \\ for Low-Resource and Multilingual Machine Translation}

\author{
	Naman Goyal\footnotemark[1], Cynthia Gao\footnotemark[1], Vishrav Chaudhary, Peng-Jen Chen, Guillaume Wenzek, \\\bf Da Ju, Sanjana Krishnan, Marc'Aurelio Ranzato\footnotemark[2], Francisco Guzm\'{a}n\footnotemark[2], Angela Fan\footnotemark[2] \footnotemark[3]\\ 
	Facebook AI Research, \footnotemark[3] LORIA\\ 
	\texttt{flores@fb.com} \\
}

\date{}

\begin{document}
\maketitle

\renewcommand*{\thefootnote}{\fnsymbol{footnote}}

\footnotetext[1]{Indicates equal contribution}
\footnotetext[2]{Indicates equal contribution}

\renewcommand*{\thefootnote}{\arabic{footnote}}

\begin{abstract}
One of the biggest challenges hindering progress in low-resource and multilingual machine translation is the lack of good evaluation benchmarks. 
Current evaluation benchmarks either lack good coverage of low-resource languages, consider only restricted domains, or are low quality because they are constructed using semi-automatic procedures. 
In this work, we introduce the \flores{} evaluation benchmark, consisting of 3001 sentences extracted from English Wikipedia and covering a variety of different topics and domains. 
These sentences have been translated in 101 languages by professional translators through a carefully controlled process. 
The resulting dataset enables better assessment of model quality on the long tail of low-resource languages, including the evaluation of many-to-many multilingual translation systems, as all translations are multilingually aligned. 
By publicly releasing such a high-quality and high-coverage dataset, we hope to foster progress in the machine translation community and beyond.
\end{abstract}
	
\section{Introduction} 
\label{intro}

Machine translation (MT) is  
one of the most successful applications in natural language processing, as exemplified by its numerous practical applications and the number of contributions on this topic at major machine learning and natural language processing venues. 
Despite recent advances in translation quality for a handful of language pairs and domains,
MT systems still perform poorly on \textit{low-resource languages}, i.e. languages without a lot of training data. In fact, many low-resource languages are not even supported by most popular translation engines.  Yet, the majority of the world's population speak low-resource languages and would benefit from improvements in translation quality on their native languages.
As a result, the field has been shifting focus towards low-resource languages.

Over the past decade, the research community has made a lot of recent progress on models for low-resource machine translation. Approaches like iterative backtranslation~\citep{sennrich2015improving}, multilingual machine translation~\cite{gmt17,tang2020ml50, fan2020beyond}, and even unsupervised machine translation~\citep{lample_emnlp2018,artetxe_emnlp2018} have shown promising results. 
Beyond modeling, a major challenge for research in low-resource machine translation is 
evaluation. 
Low-resource evaluation is critical to the scientific progress of the field, because evaluation enables proper comparison of approaches and ultimately, a better understanding of what needs further investigation and improvement.
Unfortunately,  
finding high-quality data suitable for the evaluation process is even more difficult in low-resource scenarios.

At present, there are very few benchmarks on low-resource languages. 
These often have very low coverage of low-resource languages~\citep{riza2016introduction,thu-etal-2016-introducing,guzman-etal-2019-flores,barrault-EtAl:2020:WMT1,nekoto2020participatory,ebrahimi2021americasnli,kuwanto2021low}, limiting our understanding of how well methods generalize and scale to a larger number of languages with a diversity of linguistic features. 
There are some benchmarks that have high coverage, but these are often in specific domains, like COVID-19~\cite{tico19} or religious texts~\cite{christodouloupoulos2015massively,malaviya2017learning,tiedemann2018emerging,agic-vulic-2019-jw300}; or have low quality because they are built using automatic approaches~\citep{zhang-etal-2020-improving,schwenk2019ccmatrix,schwenk-etal-2021-wikimatrix}. 
As a result, it is difficult to draw firm conclusions about research efforts on low-resource MT. 
In particular, there are even fewer benchmarks that are suitable for evaluation of many-to-many multilingual translation, as these require multi-lingual alignment (i.e. having the translation of the same sentence in multiple languages), which hampers the progress of the field despite all the recent excitement on this research direction. 
As an additional challenge, there are no established practices for how to build such benchmark. 
Working with professional translators in low-resource languages is difficult because of their scarce availability, and because it is non-trivial to check the
quality of their work~\cite{guzman-etal-2019-flores}.

We present the \flores{} benchmark, consisting of 3001 sentences sampled from English Wikipedia and professionally translated in 101 languages.
With this dataset, we make several contributions. First, we provide the community with a high-quality benchmark that has much larger breadth of topics and coverage of low resource languages than any other existing dataset (\textsection\ref{sec:atglance}). Second, \flores{} is suitable for many-to-many evaluation, meaning that it enables seamless evaluation of 10${},$100 language pairs. 
This enables the evaluation of popular multilingual MT systems as well as the evaluation of regionally-relevant language pairs like Spanish-Aymara and Vietnamese-Thai, for example. 
Third, we thoroughly document the annotation process we followed (\textsection\ref{sec:construction}), helping the community build institutional knowledge about how to construct MT datasets. Fourth, we release not only sentences with their translation but also rich meta-data that will support other kinds of evaluations and tasks, such as document level translation, multimodal translation and text classification. Fifth, we propose a BLEU metric based on sentence piece tokenization~\cite{kudo2018sentencepiece} (\textsection\ref{sec:metric}) that enables   evaluation of all languages in the set in a unified and extensible framework. Finally, we publicly release both data and baselines used in our experiments (\textsection\ref{sec:experiments}), to foster research in low-resource machine translation and related areas. 

This paper is organized as follows: In Section~\ref{sec:related_work}, we describe related work to construct evaluation benchmarks in machine translation. 
In Section~\ref{sec:construction}, we detail the construction process of \flores{}, 
from sourcing sentences to translate to defining the translation workflow.
Section~\ref{sec:atglance} gives a detailed overview of the sentences, languages, and quality of \flores{}. 
In Section~\ref{sec:metric}, we describe our proposed SentencePiece BLEU metric which unifies and simplifies evaluation. 
Section~\ref{sec:experiments} uses \flores{} to evaluate various public translation models, and breaks down model performance by amount of training data, domain, sentence length, and language family. We present our conclusions in Section~\ref{sec:conclusion}.

\begin{figure*}[t]
  \centering
  \includegraphics[width=0.9\linewidth]{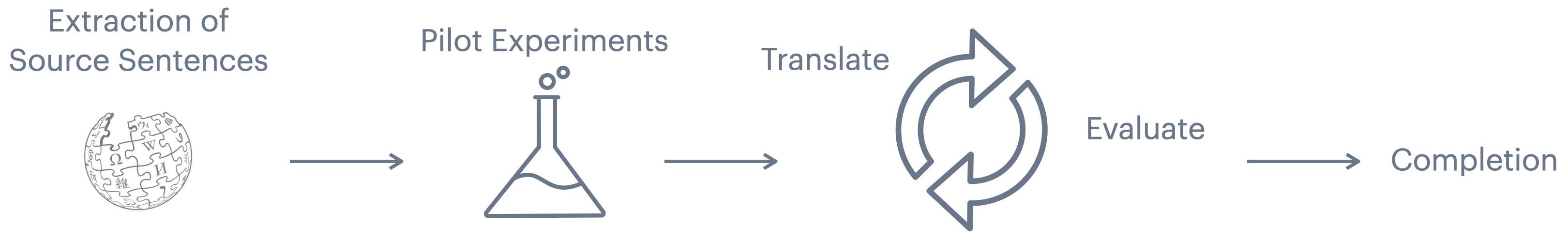}
  \caption{\textbf{Dataset Construction}. The workflow used to construct \flores{} has three
    phases: \textbf{(1)} sourcing sentences to translate from English Wikipedia, \textbf{(2)}
    designing pilot studies to define efficient and effective translation and evaluation
    processes, \textbf{(3)} launching the actual translation across all
    languages. The last stage is iterative, as translations
  may go through additional rounds of re-translation if the evaluation
  indicates that quality is insufficient; see Fig.~\ref{fig:translation_workflow} for further details.}
   \label{fig:general_outline}
 \end{figure*}

\section{Related Work} 
\label{sec:related_work}
A major challenge in machine translation, particularly as the
field shifts its focus to low-resource languages, is the
lack of availability of evaluation benchmarks. 
Much recent work has focused on the creation of training corpora~\cite{auguste2021domain,ali2021towards,adelani2021menyo,gezmu2021extended,nyoni2021low,chauhan2021monolingual} and development of models~\cite{koneru2021unsupervised,nagoudi2021indt5,aulamo2021boosting}, but evaluation is critical to being able to assess and improve translation quality.

Traditionally, the yearly Workshop on Machine Translation (WMT) and its associated shared tasks have provided standardized benchmarks and metrics to the community, fostering progress by providing means of fair comparison among various approaches. 
 Over recent years, the main translation task at WMT has challenged participants with low-resource languages, but the evaluation has been limited to a handful of languages --- for example, Latvian in 2017~\cite{ondrej2017findings}, Kazakh in 2018~\cite{rej2018findings}, Gujarati and Lithuanian in 2019~\cite{barrault2019findings}, and Inuktitut, Khmer, Pashto, and Tamil in 2020~\cite{barrault2020findings}. 
Moreover, these tasks have considered translation to and from English only, while the field has been recently focusing on large-scale multilingual models~\citep{gmt17,aharoni-etal-2019-massively,freitag-firat-2020-complete,fan2020beyond}.

To date, the largest resource of parallel data which can also be used for
evaluation purposes is OPUS~\cite{tiedemann2012parallel}, 
which is itself a collection of publicly available parallel datasets. 
While OPUS has by far the largest coverage of languages, particularly to and from English, it consists of a mixture of manually translated and mined data, which results in a large variety of datasets and domains with varying level of quality. 
For instance, OPUS contains parallel data translated by humans coming from operating system handbooks like Ubuntu, or parallel data from religious documents~\cite{liu2021usefulness} like Jehovah's Witness magazines~\citep{agic-vulic-2019-jw300} and the Bible. 
These have recently been expanded to include more languages~\cite{nicolai2021expanding} as well. 
OPUS also contains a variety of other automatically-aligned datasets, such as various versions of TED talks, which are usually of lower quality
~\citep{Ye2018WordEmbeddings,zhang-etal-2020-improving,fan2020beyond}. 
Similarly, OPUS contains large parallel datasets generated via automatic filtering and alignment methods, such as WikiMatrix~\citep{schwenk-etal-2021-wikimatrix}, ccMatrix~\cite{schwenk2019ccmatrix}, ccAligned~\cite{elkishky2020ccaligned}, and ParaCrawl~\cite{paracrawl:2019:mt_summit}, which contain noisy translations. 
While these may be utilized for training, they are clearly unsuitable for evaluation purposes due to automatic alignment. 

There are other datasets for evaluation purposes, such as Flores v1.0~\citep{guzman-etal-2019-flores}, LORELEI~\citep{strassel-tracey-2016-lorelei},  ALT~\citep{thu-etal-2016-introducing,riza2016introduction,ding2016similar} and TICO-19~\citep{tico19}, as well as datasets for specific languages such as Igbo~\cite{ezeani2020igbo} and Fon~\citep{dossou2020ffr}. These are similar to \flores{} because they focus on low-resource languages. 
However, the language coverage of these datasets is much smaller. Among these, only TICO-19 is suitable for multilingual machine translation, but its content is centered around COVID-19, unlike the much broader coverage of topics offered by \flores{}.

Lastly, the current literature in low-resource translation provides very scarce guidance in terms of best practices and methodology to construct parallel datasets and perform quality assurance. The much lower number of translators is problematic because it makes the annotation process much more susceptible to variance in the proficiency level of such annotators. 
In Flores v1.0~\citep{guzman-etal-2019-flores}, a mixture of human and automatic checks were used to filter and rework problematic translations. In TICO-19~\cite{tico19}, a two-step translation and quality assurance process was followed. Despite its technical complexity, the annotation process for benchmark sets is
in fact seldom documented in technical reports. This is still largely an uncharted territory. 
However, there are a lot of practical questions related to setting up and ensuring the quality of large-scale translation campaigns targeting low-resource languages which may have very few annotators. For example: 
What guidelines should be considered for translators and evaluators? 
What workflow is most efficient and effective? 
What automatic checks should be put in place to minimize human intervention? 
When can a dataset be declared to have reached a sufficient level of quality to be released?
In this study, we document our choices and processes in a hope to build and consolidate best practices of dataset construction for the machine translation community.

\section{Dataset Construction} 
\label{sec:construction}
The construction of \flores{} is intended to accomplish several goals: 
\Ni 
to enable the evaluation of many-to-many multilingual models, meaning the evaluation of translations from any language to any other language including very long-tail languages;
\Nii  
to enable other kinds of evaluation beyond machine translation, such as document-level translation, multi-modal translation, multilingual classification, and so on; 
\Niii 
most importantly, 
to build a high-quality evaluation benchmark. 

To achieve the above goals, the overall construction process consisted of three phases, as outlined in Fig~\ref{fig:general_outline}. 
First, we extracted sentences to translate from English Wikipedia. 
Second, we designed and ran pilot experiments to determine the translation process, and finally we launched the actual translation workflow for over 100 languages. 
In this section, we describe the process in detail. The reader who is more curious about general statistics can safely skip this section and go to Section~\ref{sec:atglance}.

\subsection{Sourcing Sentences} 
\label{sec:sourcing}

We describe how the domains and sentences in \flores{} were selected. A high-level summary of the dataset can be found in Table~\ref{tab:source_stats}.

\paragraph{Original Source.} All source sentences were extracted from multiple Wikimedia sources, as this is a repository of text that is public and freely available under permissive licensing, and covers a broad range of topics. 
Although Wikipedia is currently supported in more than 260
languages\footnote{\url{https://en.wikipedia.org/wiki/Wikipedia:Multilingual_statistics}}, several low-resource languages have relatively few articles containing well structured sentences. 
Moreover, translating a few hundred sentences for several thousand different language pairs would be infeasible, at the very least because of the lack of qualified professional translators that could read both the source and target side.

Instead, we opted to source all sentences from English Wikipedia, while considering a broad set of topics that could be of general interest regardless of the native language of the reader. 
In particular, we collected a third of the sentences from \textit{Wikinews}\footnote{\url{https://en.wikinews.org/wiki/Main_Page}}, which is a collection of international news articles, a third from \textit{Wikijunior}\footnote{\url{https://en.wikibooks.org/wiki/Wikijunior}}, which is a collection of age-appropriate nonfiction books for children from
birth to age 12, and a third from   \textit{WikiVoyage}\footnote{\url{https://en.wikivoyage.org/wiki/Main_Page}} which is a travel guide with a collection of articles about travel tips, food and destinations around the globe. 
By translating the same set of English sentences in more than hundred languages, we enable evaluation of multilingual MT with the only caveat that \textit{source} sentences not in English are produced by human translators. 
While translationese (or overly literal or awkward translations) has known idiosyncrasies~\citep{zhang-toral-2019-effect}, 
we conjecture that these effects are rather marginal when evaluating models in low-resource languages, where current MT systems produce many severe mistakes. 
We believe the benefits of many-to-many evaluation, which supports the measurement of traditionally neglected regionally-relevant pairs such as Xhosa-Zulu, Vietnamese-Thai, and Spanish-Aymara, largely outsize the risk of evaluating translationese.

\paragraph{Sentence Selection.} 
The sentence selection process consisted of selecting an article at random from each source, and then manually selecting a few (typically between 3 and 5) contiguous sentences from each article, avoiding segments with very short or malformed sentences. 
To avoid bias coming from the document structure, 
we carefully selected one paragraph per document, from either the beginning, middle or end of the article. We balanced the location selection to be equally distributed across the whole corpus --- roughly one third of paragraphs were sampled from the beginning of the article, one third from the middle, and so on. 
For each sentence, we also extracted the Wikipedia URL, topic, and noted boolean flags to indicate whether the sentence contained entities linked to other Wikipedia pages and images. 
The selection process was performed by 10 different annotators in our lab, 6 male and 4 female;  with different roles in research: researchers (scientists and engineers) and program/project managers; originally coming from different regions of the world: East Asia, South Asia, Southern Europe, Latin America and North America.
We manually labeled all sentences by a more detailed \textit{sub-topic}, one of 10 possibilities: crime, disasters, entertainment, geography, health, nature, politics, science, sports, and travel. 
Table~\ref{tab:source_stats} reports basic statistics of the originating English sentences.

Since several contiguous sentences are extracted from the same article and since we also provide the corresponding URL, we support evaluation of machine translation at the document level. 
With the additional meta-data, we also enable evaluation of multimodal machine
translation.

\begin{table}[t]
  \small
  \begin{tabular}{lrr}
    \toprule
  \multicolumn{2}{l}{Number of Sentences} & 3001 \\
  \multicolumn{2}{l}{Average Words per Sentence} & 21 \\
  \multicolumn{2}{l}{Number of Articles} & 842 \\
  \multicolumn{2}{l}{Average Number of Sentences per Article} & 3.5 \\
  \multicolumn{2}{l}{\% of Articles with Hyperlinked Entities} & 40 \\
  \multicolumn{2}{l}{\% of Articles with Images} & 66 \\
  \midrule 
  \bf Evaluation Split & \bf \# Articles & \bf \# Sentences \\
  dev & 281 & 997 \\ 
  devtest & 281 & 1012 \\ 
  test & 280 & 992 \\ 
  \midrule 
  \bf Domain & \bf \# Articles & \bf \# Sentences \\
  WikiNews & 309 & 993 \\ 
  WikiJunior & 284 & 1006 \\ 
  WikiVoyage & 249 & 1002 \\ 
  \midrule 
  \bf Sub-Topic & \bf \# Articles & \bf \# Sentences\\ 
  Crime  & 155 & 313 \\ 
  Disasters  & 27 & 65\\ 
  Entertainment  & 28 & 68 \\ 
  Geography  & 36 & 86\\ 
  Health & 27  & 67\\ 
  Nature  & 17 & 45\\  
  Politics & 171 & 341 \\ 
  Science & 154 & 325\\ 
  Sports & 154 & 162\\ 
  Travel & 505 & 1529 \\ 
  \bottomrule
\end{tabular}
\caption{\textbf{Statistics of \flores{}.} \flores{} contains 3001 sentences selected from 842 articles, divided into three splits: dev, devtest, and test. The articles are sourced from three domains, breaking down into 10 sub-topic classifications.
}
\label{tab:source_stats}
\end{table}

\subsection{Pilot Experiments}
\label{sec:pilot}

Obtaining high translation quality in low-resource languages is difficult because the translation job relies on the skill of a small set of translators.
If one translator is not perfectly fluent or uses a different local convention for that language, this could render the quality of the dataset insufficient or inconsistent for that language. 
Here, we describe the process we followed to define an efficient and high quality translation workflow. 
To this end, we report two pilot experiments we used to determine how we should proceed with the creation of this large-scale evaluation dataset.

\begin{figure*}[t]
  \centering
  \includegraphics[width=0.9\linewidth]{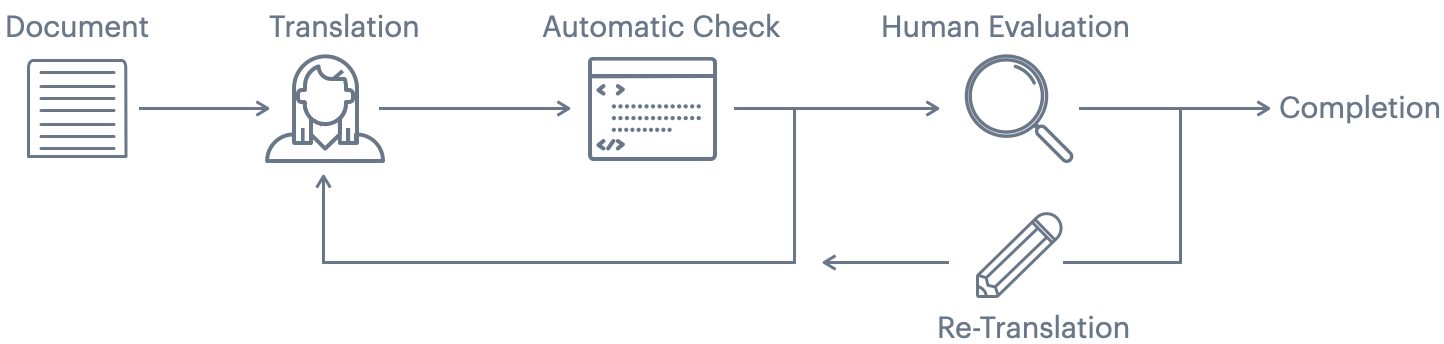}
  \caption{\textbf{Depiction of Overall Translation Workflow.} For each target
    language, sentences are translated by a translation Language Service Provider (LSP). The
    resulting translations are automatically checked. If these checks
    fail, translations are sent back for re-translation. If the
    automatic checks pass, translations are sent to another LSP
    for human evaluation. If the quality is not sufficient,
    translations are sent back to the original translation LSP for
    re-translation. Depending on the human score, the process can repeat
    for multiple rounds of human evaluation.
  }
   \label{fig:translation_workflow}
 \end{figure*}

\subsubsection{Providers and Workflows} 
To ensure the best possible level of quality for our translations, we designed two pilots aimed to determine the best workflow to follow for translating hundreds of languages. The first pilot experiment was meant to select the best translation providers for each language and the second, to determine the best translation-quality assurance workflow. 

\paragraph{Language Service Providers.} 

As a starting point, let us assume that each language can be translated by $K$ different Language Service Providers (LSPs) and that they all charge the same price for translating a sentence. 
We randomly selected 100 sentences and 8 language pairs, and assigned each language to at least two LSPs. 
We then used another LSP to evaluate all translations.

Based on human assessment of translation quality, we selected the two LSPs that produced the highest quality translations.  
We chose two translation LSPs to make our translation process not rely entirely on a single-party, while reducing the communication overhead created by working with too many external parties.

\paragraph{Translation and Quality Assurance Workflow.} 

Despite having reliable translation LSPs, we need to ensure that each translation conforms to the highest level of quality required by a benchmark. Therefore, we split our workflow into two parts: translation (which includes editing), performed by an initial LSP, and quality assurance (QA) performed by an independent LSP. After the QA process, a translation might need \textit{re-translation} or minimal editing to improve its quality.
Here, we explored the best workflow for when  \textit{re-translation} is needed. 
Assuming there are two translation LSPs, A and B, and a separate QA LSP C, we can have two possible workflows:
\Ni \textbf{A--C--B} we can have B re-translate translations flagged by C that were produced by A;
\Nii \textbf{A--C--A} an alternative and simpler workflow is to have the same LSP take care of both translation and re-translation for a given language, and to have each translation LSP handle half of the languages.

The advantage of workflow \Ni is the re-translation process is the least biased, particularly on low-resource languages where the re-translator and the translator could be the same person. On the other hand, the re-translator has less context and the workflow has higher complexity because data comes in and out of LSPs at different times, making the whole process more error prone.
 
We tested both workflows and observed negligible differences between the two workflows, and therefore, we chose workflow \Nii, i.e. {A--C--A}, with the same LSP taking care of both translation and re-translation as it is operationally simpler.

\subsubsection{Automatic Translation Quality} 
The second pilot experiment aimed to investigate how to assess translation quality automatically. 

\paragraph{Implemented Checks.} We implemented several checks to ensure the first round of translations were of acceptable quality: \Ni language identification, \Nii checking whether the translation is a copy of the source sentence; \Niii checking whether the translation has significantly different length, \Niv checking translation fluency according to a language model, \Nv and checking whether the translation is a copy of the translation produced by publicly available translation engines. 
Among all checks, we found that \Nv was the most significant issue, despite formulating clear guidelines forbidding the use of translation engines.
This is important, as we want to our translations to be as unbiased as possible. Relying on verbatim or post-edited translations from online engines would be misguiding, and give them unfair advantage when using a reference-based automatic metric for comparison.

To address the issue, we proposed a heuristic to detect and reject professional translations that are likely copies from translation engines based on their sentence-level similarity.
Moreover, when the translations of two different engines are available, we check whether a sentence is very similar to the output of a specific translation engine while being different from the output produced by other translation engines.

The heuristic is as follows: let $x$ be the translation
produced by a translation LSP, $y_A$ and $y_B$ be translations
produced by translation engine A and B; and let spBLEU($x$, $y$) be the sentence-level SentencePiece BLEU (cfr. Sec. \ref{sec:metric}) between
sentence $x$ and the reference $y$. Then, we declare a that a translation was a copy if $\mbox{spBLEU}(x, y_A) - \mbox{spBLEU}(x, y_B) > 20 \mbox{ and } \mbox{spBLEU}(x, y_A) > 50$, when the language in question is supported by both engines A and B; and if $\mbox{spBLEU}(x, y_A) > 50$ when the translation is only supported by translation engine A.\\


The values of $50$ and $20$ were based on the analysis of the distribution of scores for translations of tens of languages, where we used clustering techniques to determine the right cutoff values.

\begin{figure}[t]
  \centering
  \includegraphics[width=0.9\linewidth]{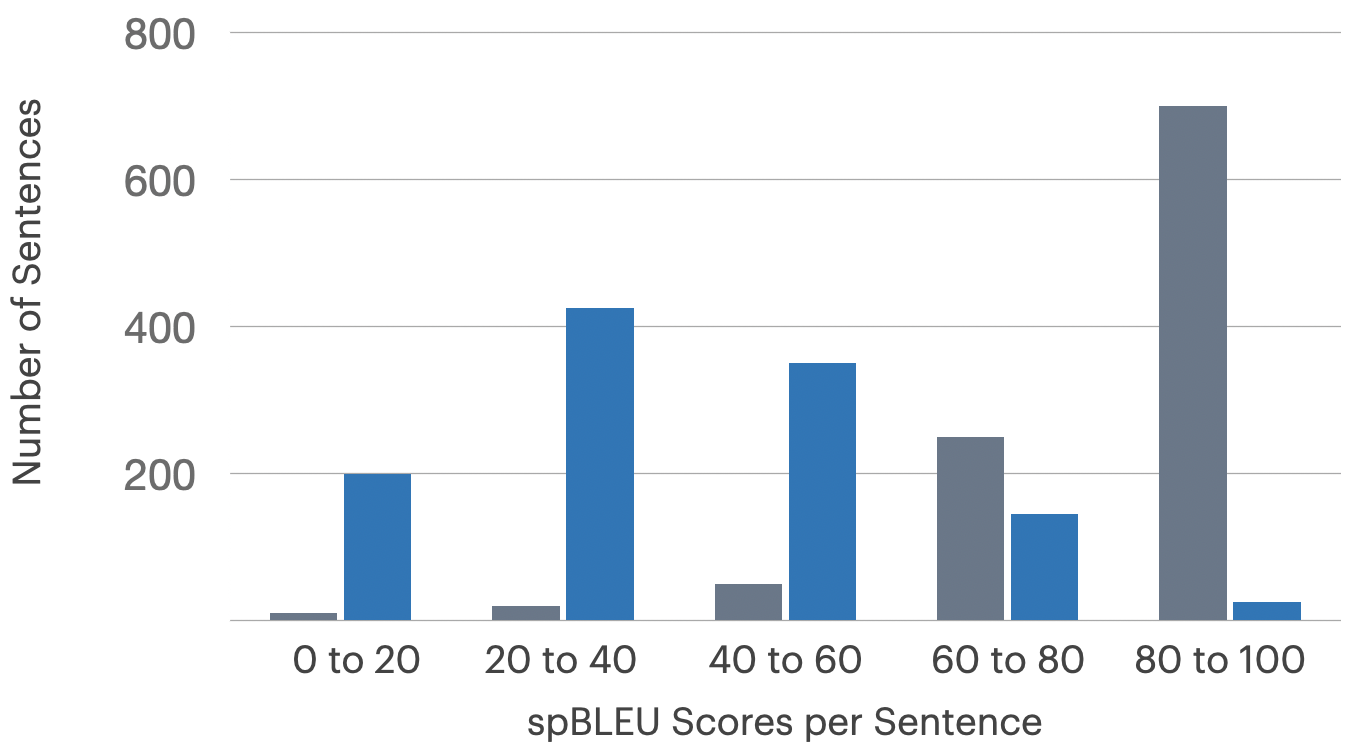}
  \caption{\textbf{Importance of Automatic Checks}. In \textit{gray}, we show the sentence-level spBLEU of a language that displays indication of copy from a commercial translation engine. A large number of sentences have very high BLEU scores, mainly in the 80 to 100 BLEU bucket. In contrast, the bars in \textit{blue} indicate a language that does not experience this issue. We discuss the spBLEU metric in greater detail in Section~\ref{sec:metric}.}
   \label{fig:google_copy}
\end{figure}

Moreover, we established that any set of translations with more than 10\% of the sentences violating the above criteria condition would need to be re-translated prior to perform any subsequent human evaluation. 
We show in Figure~\ref{fig:google_copy} an example of a language that passed this test and one which did not. 
Thanks to these automatic checks, we reduced the amount of copying from popular translation engines, streamlined the translation workflow before human evaluators assessed quality, and fully automated the process of translation, evaluation, and re-translation.
This is described in the next section.

\begin{table}[t]
  \small
  \begin{tabular}{lr}
    \toprule
  \# of Languages requiring Re-translation & 45\\ 
  Avg \# of Re-translations & 1 \\ 
  Max \# of Re-translations & 3 \\ 
  \midrule 
  Avg \# of Days to Translate 1 language & 26\\ 
  Avg \# of Days to Re-Translate & 35\\ 
  Avg \# of Days for 1 language & 61\\
  Shortest Turnaround (days) for 1 language & 31\\ 
  Longest Turnaround (days) for 1 language & 89\\ 
  \bottomrule
\end{tabular}
\caption{\textbf{Statistics of \flores{} Translation Workflow.} To ensure high quality, our translation workflow includes translation and re-translation steps. We break down the amount of re-translation required, and summarize that to complete one language, it takes on average two months.}
\label{tab:translation_stats}
\end{table}

\subsection{Translation and Evaluation} 
\label{sec:translationeval}

We describe the final workflow for collecting data for all languages in \flores{}. We start with how we decide when a language is ready to be included in \flores{}: the final translation quality score. Then, we detail the full translation process, including automatic and human quality checks.

\paragraph{Translation Quality Score.} 
How do we know if the translations are good enough to include in \flores{}, and how do we know when a language has completed translation?
Before we summarize the workflow to produce translations, we briefly discuss how we measure translation quality. 
We assess translation quality through a \textit{Translation Quality Score}, calculated per language on a 0 to 100 scale.
The translation quality score is determined based on the number of identified errors by the evaluation LSPs.
The following errors are examined: grammar, punctuation, spelling, capitalization, addition or omission of information, mistranslation, unnatural translation, untranslated text, and register. 
Each error is also associated with a severity level, between minor, major, and critical. 
Based on tallying these different error types, the overall final score is determined.
We encouraged evaluators to pay particularly high attention to unnatural translation errors. 
Based on our pilot experiments, we set the acceptable translation quality score to 90\%. 

\paragraph{Translation Workflow.} 
The overall translation workflow is depicted in Figure~\ref{fig:translation_workflow}. 
For each language, all source sentences are sent to a certain translation LSPs. 
Once sentences are translated, the data is sent to different translators within the LSP for editing and then moves on to automated quality control steps. An additional verification step is added to this specific workflow with comparison of the translated data to translations from commercial engines as previously mentioned. 
If any of the checks fail, the LSP has to re-translate until all verification is passed. 
Afterwards, translations are sent to an evaluation LSP that performs quality assessment, providing a translation quality score and constructive linguistic feedback both on the sentence and language levels. 
If the score is below the accepted threshold, translations together with the assessment report are sent back to the translation LSP for re-translation. 
If the initial score is below another certain threshold (that is associated to good translation quality), the re-translated translations are evaluated by humans one more time. 
We summarize in Table~\ref{tab:translation_stats} the overall statistics around the translation process.
We include the guidelines used for quality evaluation in the Appendix.

\section{\flores{} At a Glance} \label{sec:atglance}

In this section, we analyze \flores{}. 
We provide a high level comparison of \flores{} with existing benchmarks, then discuss the sentences, languages, and translation quality in detail.

\subsection{Comparison with Existing Benchmarks} 

We compare \flores{} with several existing benchmarks, summarized in Table~\ref{tab:high_level_comparison}. 
\flores{} combines large language coverage with topic diversity, support for many-to-many evaluation, and high quality human translations (e.g. produced with no automatic alignment). Further, \flores{} adds document-level evaluation and support multimodal translation evaluation. 

\begin{table*}[ht]
\small
\centering
\begin{tabular}{l|cccccc}
\toprule 
    & \bf \# Languages & \bf Diverse  & \bf Many to & \bf Human & \bf Document & \bf Multi \\
    & & \bf Topics & \bf  Many & \bf Translations & \bf Level & \bf modal \\
\midrule  
 \textsc{FLORES} v1~\cite{guzman-etal-2019-flores} & 2 & \textcolor{cadmiumgreen}{\cmark} & \textcolor{carnelian}{\xmark} & \textcolor{cadmiumgreen}{\cmark} &  \textcolor{carnelian}{\xmark} &  \textcolor{carnelian}{\xmark}\\
  AmericasNLI~\cite{ebrahimi2021americasnli} & 10 & \textcolor{cadmiumgreen}{\cmark} & \textcolor{cadmiumgreen}{\cmark} & \textcolor{cadmiumgreen}{\cmark} & \textcolor{carnelian}{\xmark} & \textcolor{carnelian}{\xmark} \\ 
 ALT~\cite{riza2016introduction} & 13 & \textcolor{cadmiumgreen}{\cmark} & \textcolor{cadmiumgreen}{\cmark} & \textcolor{cadmiumgreen}{\cmark} & \textcolor{carnelian}{\xmark} &  \textcolor{carnelian}{\xmark}\\
 Europarl~\cite{koehn2005europarl} & 21 & \textcolor{carnelian}{\xmark} & \textcolor{cadmiumgreen}{\cmark} & \textcolor{carnelian}{\xmark} &  \textcolor{cadmiumgreen}{\cmark} &  \textcolor{carnelian}{\xmark}\\ 
 TICO-19~\cite{tico19} & 36 & \textcolor{carnelian}{\xmark} & \textcolor{cadmiumgreen}{\cmark} & \textcolor{cadmiumgreen}{\cmark} &  \textcolor{carnelian}{\xmark} &  \textcolor{carnelian}{\xmark}\\
 OPUS-100~\cite{zhang-etal-2020-improving}& 100 & \textcolor{cadmiumgreen}{\cmark} & \textcolor{cadmiumgreen}{\cmark} & \textcolor{carnelian}{\xmark} & \textcolor{carnelian}{\xmark} &  \textcolor{carnelian}{\xmark}\\
 M2M~\cite{fan2020beyond} & 100 & \textcolor{carnelian}{\xmark} & \textcolor{cadmiumgreen}{\cmark} & \textcolor{cadmiumgreen}{\cmark}\textcolor{carnelian}{\xmark} &  \textcolor{carnelian}{\xmark}&  \textcolor{carnelian}{\xmark}\\
 \midrule 
 \bf \flores{} & 101 & \textcolor{cadmiumgreen}{\cmark} & \textcolor{cadmiumgreen}{\cmark} & \textcolor{cadmiumgreen}{\cmark} &  \textcolor{cadmiumgreen}{\cmark} & \textcolor{cadmiumgreen}{\cmark} \\
 \bottomrule 
\end{tabular}
\caption{\textbf{Comparison of Various Evaluation Benchmarks}. We compare \flores{} to a variety of popular, existing translation benchmarks, indicating language coverage, topic diversity, whether many-to-many translation is supported, if the translations are created by humans, and if the tasks of document-level translation or multimodal translation are supported.}
\label{tab:high_level_comparison}
\end{table*}

\begin{figure*}[ht]
\begin{center}
\includegraphics[width=1\linewidth]{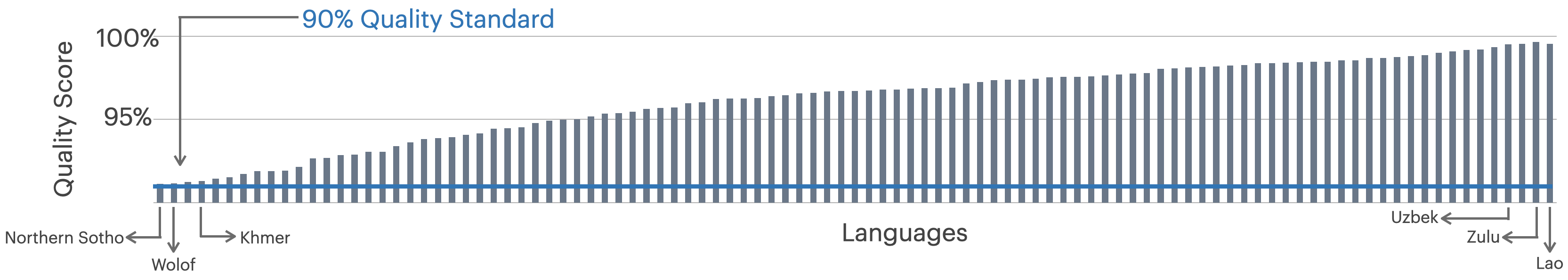}
\end{center}
\caption{\textbf{Translation Quality Score across Languages.} We require the final translation quality score to be above 90\% before the translation is of sufficient quality to include in \flores{}. We depict the score distribution for all languages in \flores{}.}
\label{fig:qa_score}
\end{figure*}

\subsection{Sentences in \flores{}} 
Table~\ref{tab:source_stats} provides an overview of \flores{}. 
The total dataset translates 3001 sentences into 101 languages. 
On average, sentences contain around 20 words. 
These sentences originate from 1,175 different articles in three domains: WikiNews, WikiJunior, and WikiVoyage. 
On average, 3 sentences are selected from each document, and then documents are divided into dev, devtest, and test sets. 
The articles are rich in metadata: 40\% of articles contain hyperlinks to other pages, and 66\% of articles contain images. 
We manually classify the content of the sentences into one of 10 broader topics, and display the distribution. 
Overall, most sentences are about world travel (sourced from WikiVoyage), though there are also a large number of sentences about science, politics, and crime. 

\subsection{Languages in \flores{}} 

We summarize all 101 languages in \flores{} and their scripts and language families in Table~\ref{tab:all_languages}. 
We note that language classification is a complex task with different classification hierarchies. 
We chose language families at a 
reasonable level of detail, i.e. fine enough such that languages can be grouped with a few other languages but not so fine that each language is in its own group. 
Overall, our selected languages cover a large percentage of people all over the world, with a large diversity of scripts and families. 
Most of these languages are spoken by millions of people, despite being considered low-resource in the research community. 

In Table~\ref{tab:all_languages}, we depict all \flores{} languages by their resource level. 
The amount of data available for a language is difficult to accurately denote, in part because quality is very important and thus, the amount of data does not necessary reflect its usefulness. 
Further, some data may be proprietary, and new datasets for new languages are actively being created by the research community.
Thus, we report the amount of data to/from English and the amount of monolingual data available in OPUS, a public repository for multilingual data.
OPUS is a heavily used resource, and itself is a collection of a large number of research datasets produced by the community over decades. 
The majority of languages have both bilingual data through English and monolingual data, though a number of languages have less than 100K sentences through English.
Many of those also have no monolingual data available, making these truly low-resource. 
Examples include Shona and Nyanja.

\begin{table*}[p]
    \small
    \centering
    \begin{tabular}{l l l l lll  }
    \toprule 
         \bf ISO 639-3 & \bf Language & \bf Family & \bf Subgrouping & \bf  Script & \bf Bitext & \bf Mono \\
         \bf  & \bf  & \bf   & \bf  & \bf   & \bf w/ En & \bf Data \\

    \midrule 
afr	&	\bf Afrikaans	&	Indo-European	&	Germanic	&	Latin	&	570K	&	26.1M	\\
amh	&	\bf Amharic	&	Afro-Asiatic	&	Afro-Asiatic	&	Ge'ez	&	339K	&	3.02M	\\
ara	&	\bf Arabic	&	Afro-Asiatic	&	Afro-Asiatic	&	Arabic	&	25.2M	&	126M	\\
hye	&	\bf Armenian	&	Indo-European	&	Other IE	&	Armenian	&	977K	&	25.4M	\\
asm	&	\bf Assamese	&	Indo-European	&	Indo-Aryan	&	Bengali	&	43.7K	&	738K	\\
ast	&	\bf Asturian	&	Indo-European	&	Romance	&	Latin	&	124K	&	---	\\
azj	&	\bf Azerbaijani	&	Turkic	&	Turkic	&	Latin	&	867K	&	41.4M	\\
bel	&	\bf Belarusian	&	Indo-European	&	Balto-Slavic	&	Cyrillic	&	42.4K	&	24M	\\
ben	&	\bf Bengali	&	Indo-European	&	Indo-Aryan	&	Bengali	&	2.16M	&	57.9M	\\
bos	&	\bf Bosnian	&	Indo-European	&	Balto-Slavic	&	Latin	&	187K	&	15.9M	\\
bul	&	\bf Bulgarian	&	Indo-European	&	Balto-Slavic	&	Cyrillic	&	10.3M	&	235M	\\
mya	&	\bf Burmese	&	Sino-Tibetan	&	 \scriptsize Sino-Tibetan+Kra-Dai 	&	Myanmar	&	283K	&	2.66M	\\
cat	&	\bf Catalan	&	Indo-European	&	Romance	&	Latin	&	5.77M	&	77.7M	\\
ceb	&	\bf Cebuano	&	Austronesian	&	Austronesian	&	Latin	&	484K	&	4.11M	\\
zho	&	\bf Chinese { \scriptsize (Simpl) }	&	Sino-Tibetan	& \scriptsize Sino-Tibetan+Kra-Dai 	&	Han	&	37.9M	&	209M	\\
zho	&	\bf Chinese { \scriptsize (Trad) }	&	Sino-Tibetan	&	\scriptsize Sino-Tibetan+Kra-Dai 	&	Han	&	37.9M	&	85.2M	\\
hrv	&	\bf Croatian	&	Indo-European	&	Balto-Slavic	&	Latin	&	42.2K	&	144M	\\
ces	&	\bf Czech	&	Indo-European	&	Balto-Slavic	&	Latin	&	23.2M	&	124M	\\
dan	&	\bf Danish	&	Indo-European	&	Germanic	&	Latin	&	10.6M	&	344M	\\
nld	&	\bf Dutch	&	Indo-European	&	Germanic	&	Latin	&	82.4M	&	230M	\\
est	&	\bf Estonian	&	Uralic	&	Uralic	&	Latin	&	4.82M	&	46M	\\
tgl	&	\bf Filipino { \scriptsize (Tagalog)}	&	Austronesian	&	Austronesian	&	Latin	&	70.6K	&	107M	\\
fin	&	\bf Finnish	&	Uralic	&	Uralic	&	Latin	&	15.2M	&	377M	\\
fra	&	\bf French	&	Indo-European	&	Romance	&	Latin	&	289M	&	428M	\\
ful	&	\bf Fula	&	Atlantic-Congo	&	\scriptsize Nilotic+Other AC 	&	Latin	&	71K	&	531K	\\
glg	&	\bf Galician	&	Indo-European	&	Romance	&	Latin	&	1.13M	&	4.22M	\\
lug	&	\bf Ganda	&	Atlantic-Congo	&	Bantu	&	Latin	&	14.4K	&	537K	\\
kat	&	\bf Georgian	&	Kartvelian	&	Other	&	Georgian	&	1.23M	&	31.7M	\\
deu	&	\bf German	&	Indo-European	&	Germanic	&	Latin	&	216M	&	417M	\\
ell	&	\bf Greek	&	Indo-European	&	Other IE	&	Greek	&	23.7M	&	201M	\\
guj	&	\bf Gujarati	&	Indo-European	&	Indo-Aryan	&	Gujarati	&	160K	&	9.41M	\\
hau	&	\bf Hausa	&	Afro-Asiatic	&	Afro-Asiatic	&	Latin	&	335K	&	5.87M	\\
heb	&	\bf Hebrew	&	Afro-Asiatic	&	Afro-Asiatic	&	Hebrew	&	6.64M	&	208M	\\
hin	&	\bf Hindi	&	Indo-European	&	Indo-Aryan	&	Devanagari	&	3.3M	&	104M	\\
hun	&	\bf Hungarian	&	Uralic	&	Uralic	&	Latin	&	16.3M	&	385M	\\
isl	&	\bf Icelandic	&	Indo-European	&	Germanic	&	Latin	&	1.17M	&	37.5M	\\
ibo	&	\bf Igbo	&	Atlantic-Congo	&	\scriptsize Nilotic+Other AC 	&	Latin	&	145K	&	693K	\\
ind	&	\bf Indonesian	&	Austronesian	&	Austronesian	&	Latin	&	39.1M	&	1.05B	\\
gle	&	\bf Irish	&	Indo-European	&	Other IE	&	Latin	&	329K	&	1.54M	\\
ita	&	\bf Italian	&	Indo-European	&	Romance	&	Latin	&	116M	&	179M	\\
jpn	&	\bf Japanese	&	Japonic	&	Other	&	 \scriptsize Han, Hiragana, Katakana	&	23.2M	&	458M	\\
jav	&	\bf Javanese	&	Austronesian	&	Austronesian	&	Latin	&	1.49M	&	24.4M	\\
kea	&	\bf Kabuverdianu	&	Indo-European	&	Romance	&	Latin	&	5.46K	&	178K	\\
kam	&	\bf Kamba	&	Atlantic-Congo	&	Bantu	&	Latin	&	50K	&	181K	\\
kan	&	\bf Kannada	&	Dravidian	&	Dravidian	&	Telugu-Kannada	&	155K	&	13.1M	\\
kaz	&	\bf Kazakh	&	Turkic	&	Turkic	&	Cyrillic	&	701K	&	35.6M	\\
khm	&	\bf Khmer	&	Austro-Asiatic	&	Austro-Asiatic	&	Khmer	&	398K	&	8.87M	\\
kor	&	\bf Korean	&	Koreanic	&	Other	&	Hangul	&	7.46M	&	390M	\\
kir	&	\bf Kyrgyz	&	Turkic	&	Turkic	&	Cyrillic	&	566K	&	2.02M	\\
lao	&	\bf Lao	&	Kra-Dai	&	 \scriptsize Sino-Tibetan+Kra-Dai  &	Lao	&	153K	&	2.47M	\\
lav	&	\bf Latvian	&	Indo-European	&	Balto-Slavic	&	Latin	&	4.8M	&	68.4M	\\
lin	&	\bf Lingala	&	Atlantic-Congo	&	Bantu	&	Latin	&	21.1K	&	336K	\\
lit	&	\bf Lithuanian	&	Indo-European	&	Balto-Slavic	&	Latin	&	6.69M	&	111M	\\
luo	&	\bf Luo	&	Nilo-Saharan	&	 \scriptsize Nilotic+Other AC 	&	Latin	&	142K	&	239K	\\
ltz	&	\bf Luxembourgish	&	Indo-European	&	Germanic	&	Latin	&	3.41M	&	---	\\
mkd	&	\bf Macedonian	&	Indo-European	&	Balto-Slavic	&	Cyrillic	&	1.13M	&	28.8M	\\
msa	&	\bf Malay	&	Austronesian	&	Austronesian	&	Latin	&	968K	&	77.5M	\\
mal	&	\bf Malayalam	&	Dravidian	&	Dravidian	&	Malayalam	&	497K	&	24.8M	\\
mlt	&	\bf Maltese	&	Afro-Asiatic	&	Afro-Asiatic	&	Latin	&	5.82M	&	---	\\
mri	&	\bf Māori	&	Austronesian	&	Austronesian	&	Latin	&	196K	&	---	\\
mar	&	\bf Marathi	&	Indo-European	&	Indo-Aryan	&	Devanagari	&	109K	&	14.4M	\\
mon	&	\bf Mongolian	&	Mongolic	&	Other	&	Cyrillic	&	555K	&	20.4M	\\
npi	&	\bf Nepali	&	Indo-European	&	Indo-Aryan	&	Devanagari	&	19.6K	&	17.9M	\\
nso	&	\bf Northern Sotho	&	Atlantic-Congo	&	Bantu	&	Latin	&	13.8K	&	612K	\\
nob	&	\bf Norwegian	&	Indo-European	&	Germanic	&	Latin	&	10.9M	&	338M	\\
     \bottomrule
    \end{tabular}
\end{table*}

\begin{table*}[t]
    \small
    \centering
    \begin{tabular}{l l l l lll  }
    \toprule 
         \bf ISO 639-3 & \bf Language & \bf Family & \bf Subgrouping & \bf  Script & \bf Bitext & \bf Mono \\
         \bf  & \bf  & \bf   & \bf  & \bf   & \bf w/ En & \bf Data \\
    \midrule 
nya	&	\bf Nyanja	&	Atlantic-Congo	&	Bantu	&	Latin	&	932K	&	---	\\
oci	&	\bf Occitan	&	Indo-European	&	Romance	&	Latin	&	5.11K	&	---	\\
ory	&	\bf Oriya	&	Indo-European	&	Indo-Aryan	&	Oriya	&	5K	&	2.47M	\\
orm	&	\bf Oromo	&	Afro-Asiatic	&	Afro-Asiatic	&	Latin	&	162K	&	752K	\\
pus	&	\bf Pashto	&	Indo-European	&	Indo-Aryan	&	Perso-Arabic	&	293K	&	12M	\\
fas	&	\bf Persian	&	Indo-European	&	Indo-Aryan	&	Perso-Arabic	&	6.63M	&	611M	\\
pol	&	\bf Polish	&	Indo-European	&	Balto-Slavic	&	Latin	&	40.9M	&	256M	\\
por	&	\bf Portuguese { \scriptsize (Brazil) }	&	Indo-European	&	Romance	&	Latin	&	137M	&	340M	\\
pan	&	\bf Punjabi	&	Indo-European	&	Indo-Aryan	&	Gurmukhi	&	142K	&	5.02M	\\
ron	&	\bf Romanian	&	Indo-European	&	Romance	&	Latin	&	31.9M	&	391M	\\
rus	&	\bf Russian	&	Indo-European	&	Balto-Slavic	&	Cyrillic	&	127M	&	849M	\\
srp	&	\bf Serbian	&	Indo-European	&	Balto-Slavic	&	Cyrillic	&	7.01M	&	35.7M	\\
sna	&	\bf Shona	&	Atlantic-Congo	&	Bantu	&	Latin	&	877K	&	---	\\
snd	&	\bf Sindhi	&	Indo-European	&	Indo-Aryan	&	Perso-Arabic	&	21.8K	&	314K	\\
slk	&	\bf Slovak	&	Indo-European	&	Balto-Slavic	&	Latin	&	10.5M	&	174M	\\
slv	&	\bf Slovenian	&	Indo-European	&	Balto-Slavic	&	Latin	&	5.42M	&	74.7M	\\
som	&	\bf Somali	&	Afro-Asiatic	&	Afro-Asiatic	&	Latin	&	358K	&	14.1M	\\
ckb	&	\bf Sorani Kurdish	&	Indo-European	&	Indo-Aryan	&	Arabic	&	305K	&	7.98M	\\
spa	&	\bf Spanish { \scriptsize (Latin America) }	&	Indo-European	&	Romance	&	Latin	&	315M	&	379M	\\
swh	&	\bf Swahili	&	Atlantic-Congo	&	Bantu	&	Latin	&	349K	&	35.8M	\\
swe	&	\bf Swedish	&	Indo-European	&	Germanic	&	Latin	&	54.8M	&	580M	\\
tgk	&	\bf Tajik	&	Indo-European	&	Indo-Aryan	&	Cyrillic	&	544K	&	---	\\
tam	&	\bf Tamil	&	Dravidian	&	Dravidian	&	Tamil	&	992K	&	68.2M	\\
tel	&	\bf Telugu	&	Dravidian	&	Dravidian	&	{ \scriptsize Telugu-Kannada }	&	381K	&	17.2M	\\
tha	&	\bf Thai	&	Kra-Dai	&	{ \scriptsize Sino-Tibetan+Kra-Dai }	&	Thai	&	10.6M	&	319M	\\
tur	&	\bf Turkish	&	Turkic	&	Turkic	&	Latin	&	41.2M	&	128M	\\
ukr	&	\bf Ukrainian	&	Indo-European	&	Balto-Slavic	&	Cyrillic	&	5.44M	&	357M	\\
umb	&	\bf Umbundu	&	Atlantic-Congo	&	Bantu	&	Latin	&	217K	&	142K	\\
urd	&	\bf Urdu	&	Indo-European	&	Indo-Aryan	&	Perso-Arabic	&	630K	&	28M	\\
uzb	&	\bf Uzbek	&	Turkic	&	Turkic	&	Latin	&	---	&	7.54M	\\
vie	&	\bf Vietnamese	&	Austro-Asiatic	&	Austro-Asiatic	&	Latin	&	32.1M	&	992M	\\
cym	&	\bf Welsh	&	Indo-European	&	Other IE	&	Latin	&	826K	&	12.7M	\\
wol	&	\bf Wolof	&	Atlantic-Congo	&	Nilotic+Other AC	&	Latin	&	86.9K	&	676K	\\
xho	&	\bf Xhosa	&	Atlantic-Congo	&	Bantu	&	Latin	&	130K	&	995K	\\
yor	&	\bf Yoruba	&	Atlantic-Congo	&	Nilotic+Other AC	&	Latin	&	171K	&	1.59M	\\
zul	&	\bf Zulu	&	Atlantic-Congo	&	Bantu	&	Latin	&	123K	&	994K	\\
    \bottomrule 
    \end{tabular}
    \caption{\textbf{101 Languages in \flores{}.} We include the ISO 639-3 code, the language family, and script. Next to each language family, we include more fine-grained subgrouping information. We also include the amount of resources
  available in OPUS at the time this report was written. The parallel datasets
  were used to train the baseline described in \textsection\ref{sec:metric}, the
  monolingual datasets were only used to calculate SentencePiece,
  see Section~\textsection\ref{sec:metric}. }
    \label{tab:all_languages}
\end{table*}

\subsection{Translation Quality} 
The translation quality score across all languages is depicted in Figure~\ref{fig:qa_score}. 
All 101 languages in \flores{} meet our initial threshold of 90\% quality based on human evaluation. 
Note that several languages were considered beyond our set of 101, but were unable to meet the bar after rounds of re-translation.
Overall, about 50\% of languages have fairly high quality (above 95\%), with few near the 90\% threshold boundary. 
Even low-resource languages like Lao and Zulu can score well on the quality metric. 

We breakdown the main translation errors observed based on the quality assessments and re-translations. 
The largest error category across all languages was \textit{mistranslation}, a broad error category that generally notes that the source text was not translated faithfully and the translation has rendered an incorrect meaning in the target language. Examples of mistranslation include (but not limited to) incorrect interpretation of the source text, literal translations and mistranslations of phrasal verbs and lack of disambiguation of ambiguous terms.
For example, writing \textit{...recommends hand washing \textbf{with} hand sanitizer rubs} instead of \textit{...recommends hand washing \textbf{over} hand sanitizer rubs} would represent a mistranslation. Error categories with few errors include register, grammar, and punctuation.

We also examined if certain domains are more difficult to translate than others. 
Within a language, we did identify variation in the percentage of errors contributed by domain (often one domain could contribute up to 10\% more errors than the others), but across languages, there was no clear trend.
Overall, it appears that all domains are challenging to translate for human translators.

\section{Metric: SentencePiece BLEU} 
\label{sec:metric}

How do we evaluate the performance of translation models at the scale of 101 languages? In this section, we propose the \textit{SentencePiece BLEU} metric and analyze its performance across languages compared to various alternatives. 

\subsection{Motivation}

Automatic evaluation of translation quality is an active field. Each year, the WMT Metrics shared task seeks to determine the automatic metric that better correlates with human evaluations \cite{mathur-EtAl:2020:WMT}. 
While many metrics have been proposed through the years, the analysis has only included a handful of low-resource languages. 
Further, despite the progress in automatic metrics, the common practice is to use BLEU~\citep{bleu} when reporting results. 
Unfortunately, using BLEU directly is suboptimal, as it relies on n-gram overlap which is heavily dependent on the particular tokenization used, i.e. 
tokenizing more aggressively can artificially raise the score and make it difficult to compare across reported results. 

The challenge of making BLEU comparable by using equivalent tokenization schemes has been challenging for the translation community and has been partially addressed by  \texttt{sacrebleu}~\cite{post2018call}.
Previous standards usually leverage the \texttt{mosestokenizer}\footnote{\url{https://github.com/moses-smt/mosesdecoder/blob/master/scripts/tokenizer/tokenizer.perl}}, which is the default tool in \texttt{sacrebleu}. 
However, for many languages, these existing tools and tokenizers are not sufficient. 

For example, \texttt{mosestokenizer} supports a limited number of languages (often activated with the \texttt{--tok} flag). 
While its default tokenization rules might operate reasonably for European languages, they do not extend to global support.
For example, white-space tokenization is insufficient for some languages like Burmese or Khmer, which do not segment words with white space. Other languages like Arabic are morphologically rich, which has incentivized the creation of BLEU variants~\cite{bouamor-etal-2014-human}.
To further complicate matters, some languages like Hindi and Japanese already have custom tokenizers that are used when computing BLEU, although these appear scattered in various publication footnotes, while for others, no such special tokenizers have been developed yet.
Further, developing tokenizers for each language of interest is a challenging effort~\cite{dossou2021crowdsourced,li2021finding} that is difficult to scale.

Ideally, we would like an automatic evaluation process that is robust, simple and that can be applied to any language without the need to specify any particular tokenizer, as this will make it easier for researchers to compare against each other. 
We would like our automatic evaluation to also support future languages --- as translation quality continues to improve, the community will naturally produce models for more and more languages. 

\subsection{SentencePiece BLEU} 

Towards this goal, we have trained a SentencePiece (SPM) tokenizer~\citep{kudo2018sentencepiece} with 256,000 tokens using monolingual data~\citep{conneau2020unsupervised, wenzek2019ccnet} from all the \flores{} languages. 
SPM is a system that learns subword units based on training data, and does not require tokenization.
The logic is not dependent on language, as the system treats all sentences as sequences of Unicode. 
Given the large amount of multilingual data and the large number of languages, this essentially provides a \textit{universal} tokenizer, that can operate on any language. 

\paragraph{Training SPM.} 
One challenge is that the amount of monolingual data available for different languages is not the same --- an effect that is extreme when considering low-resource languages. 
Languages with small quantities of data may not have the same level of coverage in subword units, or an insufficient quantity of sentences to represent a diverse enough set of content. 
To address this, we train our SPM model with temperature upsampling similar to \cite{conneau2020unsupervised}, so that low-resource languages are represented. 
In the future if a new language is added to \flores{} and this tokenizer does not support its script, we can easily add new tokens to encode it as desired.

\paragraph{Computing spBLEU.} Given this SPM-tokenizer, we compute BLEU by tokenizing the system output and the reference, and then calculate BLEU in the space of sentence-pieces. We dub this metric as \textit{sentence-piece BLEU}, and we denote it as spBLEU. It is integrated into \texttt{sacrebleu} for ease of use\footnote{\url{https://github.com/ngoyal2707/sacrebleu/tree/adding_spm_tokenized_bleu}} as the \texttt{spm} tokenizer.

\begin{table}
\small
\centering
\begin{tabular}{l cc}
\toprule 
    \bf Lang &  \bf Correlation & \bf Correlation \\ 
    & \bf spBLEU v. BLEU & \bf char-BLEU v. BLEU \\
 \midrule  
 French  & 0.99 & 0.99 \\ 
 Italian & 0.99 & 0.99 \\ 
 Spanish & 0.99 & 0.99 \\ 
 \midrule 
 Hindi & 0.99 & 0.99 \\ 
 Tamil & 0.41 & 0.31 \\ 
 Chinese & 0.99 & 0.75 \\ 
 \bottomrule 
\end{tabular}
\caption{\textbf{Spearman Correlation of spBLEU, BLEU, and char-BLEU}. We evaluated on three sets of languages (En-XX). Models evaluated are derived from our baselines (discussed in Section~\ref{sec:experiments}). In the top section, we evaluate languages that often use the standard \texttt{mosestokenizer}. In the bottom section, we evaluate languages that have their own custom tokenization.}
\label{tab:metrics_comparison}
\end{table}

\subsection{Experiments and Analysis}
We want to validate the spBLEU metric, to \textbf{(1)} see that it trends with the standard BLEU metric on languages where \texttt{mosestokenizer} is often used as the default, \textbf{(2)} for languages where custom tokenizers are currently used, see that spBLEU correlates with custom-tokenizer-BLEU more strongly than alternatives such as character-level BLEU, and finally \textbf{(3)} verify that spBLEU can be used for model selection purposes.

\begin{table*}[ht]
\small
\centering
\begin{tabular}{l cc cc cc}
\toprule 
     &  \multicolumn{2}{c}{\bf spBLEU v. Human Eval} &  \multicolumn{2}{c}{\bf spBLEU v.  BLEU} & \multicolumn{2}{c}{\bf Human Eval v. BLEU}\\
     \cmidrule(r){2-3}\cmidrule(lr){4-5}\cmidrule(l){6-7}
    \bf Language &  Kendall $\tau$ & Same Best Model  &  Kendall $\tau$ & Same Best Model  &  Kendall $\tau$ & Same Best Model\\
 \midrule  
 Pashto & 1.0 & \textcolor{cadmiumgreen}{\cmark} & 1.0 & \textcolor{cadmiumgreen}{\cmark}  & 1.0 & \textcolor{cadmiumgreen}{\cmark}\\ 
 Russian & 0.71 & \textcolor{cadmiumgreen}{\cmark} & 0.52 & \textcolor{cadmiumgreen}{\cmark} & 0.80 & \textcolor{cadmiumgreen}{\cmark} \\ 
 Chinese & 0.90 & \textcolor{carnelian}{\xmark} & 0.71  & \textcolor{cadmiumgreen}{\cmark} & 0.80 & \textcolor{carnelian}{\xmark} \\ 
 \bottomrule 
\end{tabular}
\caption{\textbf{spBLEU Compared to Human Evaluation and BLEU ranking}. We analyze translation into Pashto, Russian, and Chinese. We indicate the Kendall $\tau$ between spBLEU and Human Eval, spBLEU and BLEU, and Human Eval and BLEU, as well as if the different metrics result in the selection of the same best model.}
\label{tab:spbleu_analysis_2}
\end{table*}

\paragraph{spBLEU correlates with BLEU.} 
First, we examine if spBLEU has strong positive correlation with BLEU across various languages where the \texttt{mosestokenizer} is widely used by default.
We examine Spanish, Italian, and French.
As shown in Table~\ref{tab:metrics_comparison} (top), we find that spBLEU correlates very well with BLEU on these languages. 

\paragraph{spBLEU is better than char-BLEU when custom tokenization is needed.} 
Next, we examine the performance of spBLEU on languages where custom tokenizers are often used, or special rules are written for tokenization. 
We look at three languages: Chinese, Hindi, and Tamil. 
Chinese is supported by \texttt{mosetokenizer} with special rules. 
Hindi and Tamil have a popularly used tokenizer in the community from IndicNLP\footnote{\url{https://anoopkunchukuttan.github.io/indic_nlp_library/}}.
While these language-specific tokenizers are excellent, the challenge of scale and comparability exists: there are often different competing tokenizers
and tokenizers need to be developed for each language we want to evaluate.
A possible alternative is instead to eliminate tokenization, and evaluate characters directly and compute character-level BLEU (char-BLEU). 

We examine if spBLEU is a better alternative to char-BLEU for languages that currently use special tokenizers.
As shown in Table~\ref{tab:metrics_comparison} (bottom), spBLEU correlates more strongly with custom tokenizer BLEU compared to the correlation between char-BLEU and custom tokenizer BLEU. 
While the development of custom tokenizers for specific languages produces much more accurate tokenization, spBLEU is a good alternative for comparability and scalability across a large number of languages.

\begin{table*}[t]
\small
\centering
\begin{tabular}{lccccc}
\toprule 
    & \bf Very Low & \bf Low  & \bf Medium & \bf High & \bf Avg \\\cmidrule{2-5}
    & < 100K & (100K, 1M)  &  (1M, 100M) &  > 100M &  \\
    \midrule 
    \bf Num Languages & 15 & 40 & 38 & 6 & \\ 
\midrule  
\bf Very Low & 1.60 & 2.29 & 6.98 & 9.14 & 5.00 \\
\bf Low & 2.02 & 2.74 & 8.48 & 10.30 & 5.89 \\
\bf Medium & 3.79 & 5.38 & 19.13 & 23.43 & 12.93 \\
\bf High & 4.29 & 5.83 & 21.70 & 27.32 & 14.79 \\
\midrule
\bf Avg & 2.93 & 4.06 & 14.07 & 17.55 \\
 \bottomrule 
\end{tabular}
\caption{\textbf{Many-to-Many Performance by available Bitext data through English}. We show spBLEU on devtest of \flores{} for M2M-124 615M parameter model. We group  languages into 4 bins. spBLEU is worse for low-resource languages compared to high resource languages, and translating into low-resource languages is harder than translating out of a low-resource language.}
\label{tab:many2many_bleu_resources}
\end{table*}

\paragraph{spBLEU has similar performance as BLEU for model selection.} 
Next, we turn to verifying that spBLEU can be used compare the quality of models for model selection purposes.
This is important, as oftentimes automatic metrics are used in an outer loop of the training process, to select various hyper-parameters, such as model size, dropout rate, learning rate and so on. 
Thus, we replicate the selection of various models using spBLEU instead of BLEU, experimenting on three language directions: English to Pashto, English to Russian, and English to Chinese.
We choose these languages because they were part of WMT2020 human evaluations, and thus we know the \textit{ground-truth} ranking. 
We evaluate five models for Pashto, eight for Russian, and seven for Chinese. 
We focus on evaluation of models in directions out of English, as \texttt{mosestokenizer} works well on English.

For each of the three language directions, we compute the spBLEU and BLEU with language-specific tokenizers between different systems and the reference translation. 
Overall results are shown in Table~\ref{tab:spbleu_analysis_2}. 
We first compare if the ranking of systems produced by spBLEU matches that of systems ranked by human evaluation and BLEU. 
We calculate exact match ranking accuracy using the Kendall $\tau$ coefficient, which ranges between -1 and 1.  
The ranking of systems by spBLEU matches human evaluation and BLEU perfectly for Pashto, and has strong correlation with both human evaluation and BLEU for Chinese and Russian. 

However, the exact ranking may not be the most important part of a metric. 
Oftentimes, we want to use automatic metrics to understand which model improvement is the most effective --- for example, which model to submit to WMT?
Thus, it is important to check whether the best scoring model according to spBLEU matches the best scoring model according to BLEU.
We find that spBLEU and BLEU indeed select the \textit{same} best model on all three languages. 
Note that BLEU has no guarantee of selecting the same model as human evaluators\footnote{There are a number of ways that BLEU score can be improved that may not have an effect on human evaluation. Punctuation normalization is an example.}. 

\begin{figure}[t]
  \centering
  \includegraphics[width=1\linewidth]{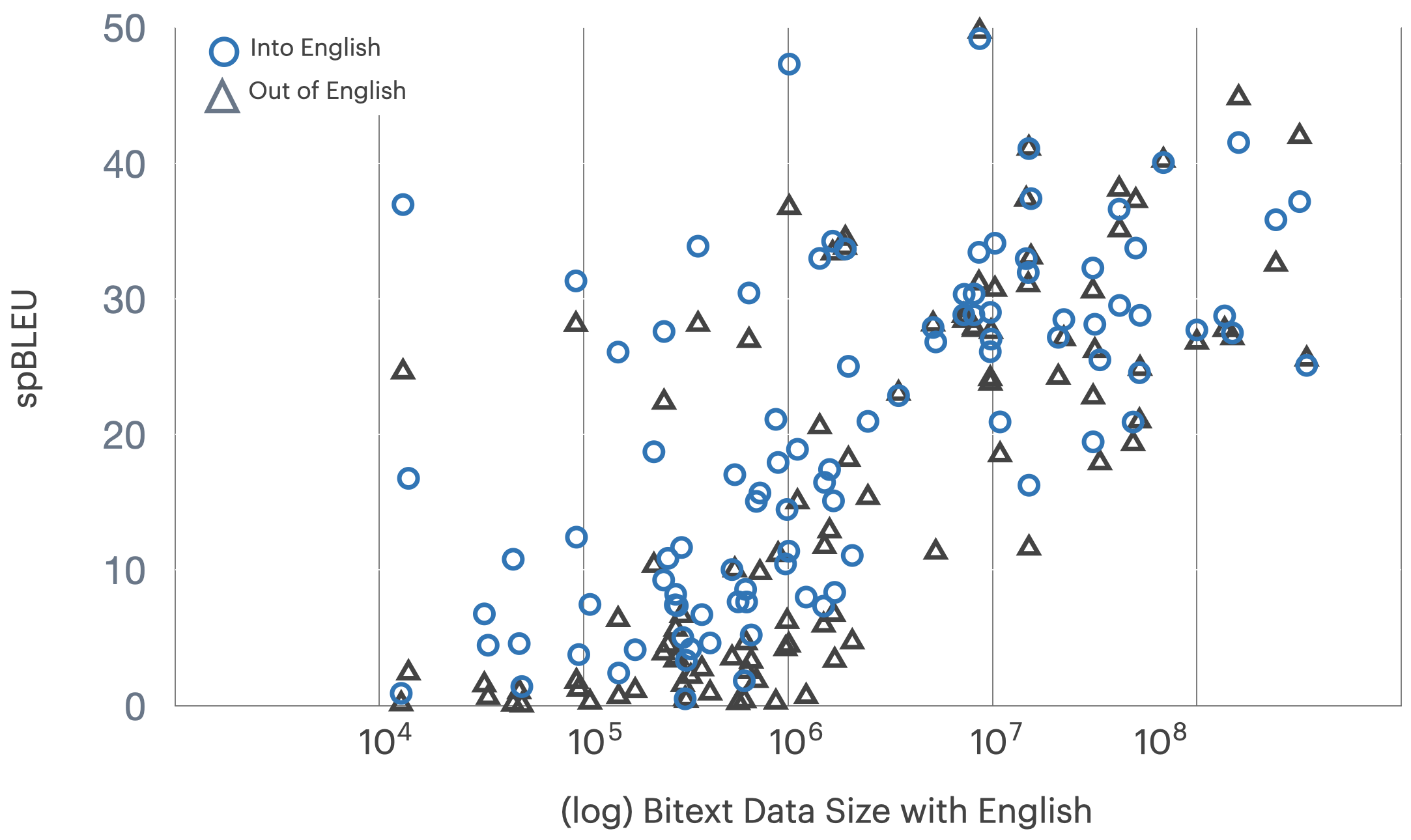}
  \caption{\textbf{spBLEU for directions in and out of English.} We compare performance against amount of available bitext data. For the same amount of data, translation into English is often stronger.}
   \label{fig:res_vs_to_from_en_bleu} 
 \end{figure}

\paragraph{Takeaway.} 
Overall, we conclude that spBLEU functions fairly similarly to BLEU, especially on languages that usually default to the \texttt{mosestokenizer}. 
On languages that use custom tokenization, spBLEU correlates more strongly with BLEU than other alternatives, such as char-BLEU. 
Further, spBLEU often produces a very similar ranking of models and selects the same best model as BLEU and human evaluation. 

For the vast majority of languages without custom tokenizers, spBLEU provides the ability to quantify performance in the same way, with one model. 
We believe that having a single model to perform tokenization will help the research community to make progress on low-resource research, while opening the door for improved versions of spBLEU that treat low-resource languages more fairly.
In the subsequent rest of the work, we use spBLEU to evaluate model performance.

\section{Evaluating Baselines on \flores{}} 
\label{sec:experiments}

In this section, we present evaluation of various models on \flores{}. We describe the dev, devtest, and test splits and how we intend them to be used. We then analyze the performance of a many-to-many model based on~\citet{fan2020beyond} and break down performance by resource level, sentence length, and language family. Finally, we compare various model variants. 

\subsection{Data Splits}

\flores{} is divided into three splits: dev, devtest, and test. 
Unless otherwise stated, we report results on the devtest portion of \flores{}. 
The dev set is meant to be used for hyper-parameter tuning. 
The devtest is meant to be used for testing purpose during the development phase. 
The test set will not be released, but will be available via a publicly available evaluation server, while the dev and devtest are publicly downloadable\footnote{\url{https://dl.fbaipublicfiles.com/flores101/dataset/flores101_dataset.tar.gz}}. 
Through the evaluation server, the test set can be used by various evaluation campaigns, such as the WMT 2021 Large-Scale Multilingual Task\footnote{\url{http://statmt.org/wmt21/large-scale-multilingual-translation-task.html}}.
The primary motivation for keeping the test set available only through an evaluation server is to guarantee equivalent assessment of models and reduce overfitting to the test set.
Further, as the dataset is many-to-many, if the source sentences are released, the target sentences would also be released. 

\begin{table*}[t]
\small
\centering
\begin{tabular}{l|ccc|c}
\toprule 
    & \bf Short & \bf Medium  & \bf Long & \bf Avg \\
    &  <= 15 words &  (15, 25) words &  > 25 words &  \\
    \midrule 
    \bf Num Sentences & 200 & 550 & 250 & \\ 
\midrule
\bf $\rightarrow$ English & 19.11 & 20.42 & 20.47 & 20.00 \\
\bf English $\rightarrow$ & 16.31 & 16.60 & 16.06 & 16.32 \\
\midrule
\bf $\rightarrow$ Chinese & 10.44 & 10.52 & 10.01 & 10.32 \\
\bf Chinese $\rightarrow$ & 9.81 & 10.21 & 9.34 & 9.79 \\
\midrule
\bf $\rightarrow$ Spanish & 13.17 & 14.08 & 14.19 & 13.81 \\
\bf Spanish $\rightarrow$ & 10.69 & 11.06 & 10.97 & 10.91 \\
\midrule
\bf $\rightarrow$ Hindi & 13.58 & 13.82 & 14.73 & 14.04 \\
\bf Hindi $\rightarrow$ & 10.42 & 10.75 & 10.54 & 10.57 \\
\midrule
\bf $\rightarrow$ Arabic & 6.83 & 7.97 & 8.84 & 7.88 \\
\bf Arabic $\rightarrow$ & 9.31 & 10.02 & 9.93 & 9.75 \\
\midrule
\bf Many-to-Many & 7.71 & 8.16 & 8.02 & \\
 \bottomrule 
\end{tabular}
\caption{\textbf{Many-to-Many Performance by Sentence Length}. We show spBLEU on devtest of \flores{} for M2M-124 615M parameter model. We analyze if sentence length has an effect on performance across directions in English, Chinese, Spanish, Hindi, Arabic, and when averaging across all language  directions. We find that length does not have a strong effect on performance.}
\label{tab:many2many_bleu_sentence_length}
\end{table*}

\subsection{Baselines} 

We evaluate on different models to provide baselines for researchers who may be interested in the performance of certain directions, and to understand which languages and directions need substantial research improvements. 

\begin{itemize}
\itemsep0em 
  \item \textbf{M2M-124}: \citet{fan2020beyond} trained the M2M-100 multilingual model by  extending large-scale data mining to create training data for language pairs not going through English.
  The original M2M-100 model does not have full coverage of the languages in \flores{}.
  We extended their mined data with data from OPUS for the \flores{} languages not present in mined data,  extending to 124 total languages.
    Note that for the additional languages added, OPUS does not contain a large quantity of data, and the OPUS data is rather noisy; see Table~\ref{tab:all_languages} for further details. 
  On this parallel data, we trained two different sizes of models, namely a model with 615M and one with 175M parameters. Unless otherwise stated, we will report results using the 615M parameter model; this is our default throughout the rest of this paper.  
    \item \textbf{OPUS-100}: \citet{zhang-etal-2020-improving} trained multilingual machine translation models on an English-centric OPUS training dataset with language-aware layers and random online backtranslation (RoBT). 
  We evaluate the 24-layer model with backtranslation (dubbed \texttt{Ours + 24 layer + RoBT} in their work\footnote{\url{https://github.com/bzhangGo/zero/tree/master/docs/multilingual_laln_lalt}}) with 254M parameters. 
    \item \textbf{Models open-sourced by Masakhane}: The Maskhane Participatory Research effort, focusing on Natural Language Processing for African languages, has developed and open-sourced for the community various machine translation models~\cite{nekoto2020participatory,abbott2019benchmarking,abbott-martinus-2019-benchmarking}. We evaluate models from English to six languages\footnote{\url{https://github.com/masakhane-io/masakhane-mt}}: Yoruba, Zulu, Swahili, Shona, Nyanja, and Luo. 

\end{itemize}

\subsection{Generation}

We generate from all models with beam size 5, setting the max generation length to 200. 
Given the large number of directions covered in \flores{}, we do not tune the beam size, length penalty, or minimum/maximum generation length.

\subsection{Results}
In this section, we report the results of the evaluation of the baseline approaches described above on the \flores{} devtest using spBLEU.

\subsubsection{Findings From Evaluation on All Directions} 

\begin{landscape}
 \begin{figure}
  \centering
  \includegraphics{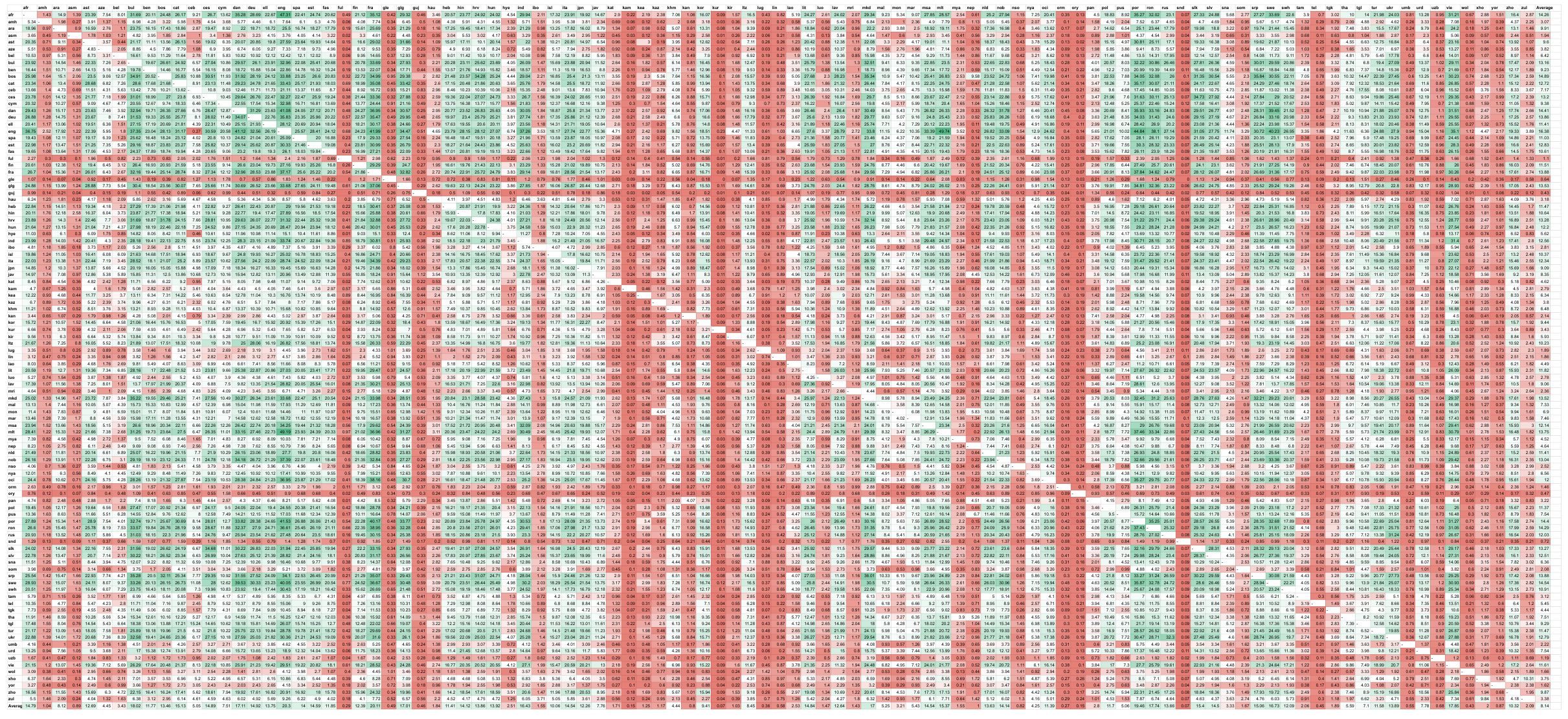}
    \caption{spBLEU of the M2M MMT model on all the language pairs of
      \flores{} dev-test set. Cell (i,j) reports spBLEU for
      translating from language i to language j. Therefore, each
      column shows spBLEU for translating in the same target language
      but using various source languages. Vice versa, each row shows
      spBLEU for translating into various target languages when
      starting from the same source language.}
    \label{fig:100x100}
 \end{figure}
\end{landscape}

\begin{landscape}
 \begin{figure}
  \centering
    \includegraphics{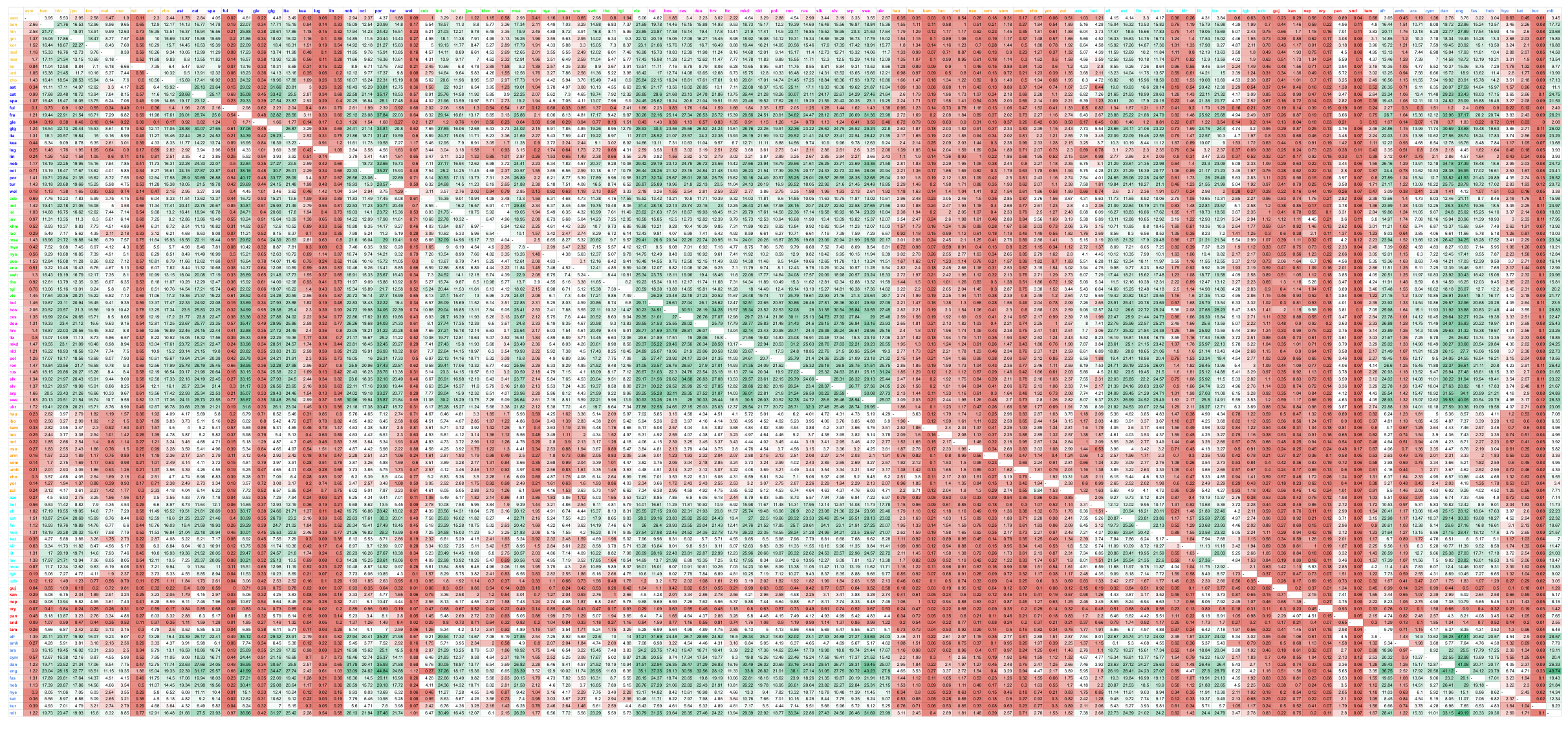}
    \caption{This is the same data as in Fig.~\ref{fig:100x100},
      except that rows and columns have been organized according to
      spectral clustering with 8 clusters (identified by the color,
      see first row and first column).}
    \label{fig:100x100sorted_nc8_crop}
 \end{figure}
\end{landscape}

\paragraph{All Directions.} We evaluated our M2M-124 model with 615M parameters on all language pairs and report spBLEU scores in Figure~\ref{fig:100x100}.
In Figure~\ref{fig:100x100}, the languages are organized alphabetically by language code, while in Figure~\ref{fig:100x100sorted_nc8_crop}, the rows and columns have been organized via spectral clustering.
The spBLEU metric scores are used as affinity scores between each pair of languages.
This produces clusters ordered by easiness to translate. 

\begin{table*}[t]
\small
\centering
\begin{tabular}{l|ccc|c}
\toprule 
    & \bf WikiNews & \bf WikiJunior  & \bf WikiVoyage & \bf Avg \\
   \bf Num Sentences & 993 & 1006 & 1002 & \\ 
\midrule
\bf English $\leftarrow$ & 20.64 & 20.67 & 19.41 & 20.24 \\
\bf English $\rightarrow$ & 16.85 & 16.67 & 15.48 & 16.33 \\
\midrule
\bf Chinese $\leftarrow$ & 11.57 & 9.66 & 9.55 & 10.26 \\
\bf Chinese $\rightarrow$ & 10.02 & 9.93 & 9.57 & 9.84 \\
\midrule
\bf Spanish $\leftarrow$ & 14.91 & 13.80 & 13.23 & 13.98 \\
\bf Spanish $\rightarrow$ & 11.67 & 10.96 & 10.37 & 11.00 \\
\midrule
\bf Hindi $\leftarrow$ & 14.33 & 14.15 & 13.84 & 14.11 \\
\bf Hindi $\rightarrow$ & 10.88 & 10.86 & 10.11 & 10.62 \\
\midrule
\bf Arabic $\leftarrow$ & 8.39 & 8.23 & 7.74 & 8.12 \\
\bf Arabic $\rightarrow$ & 9.81 & 10.31 & 9.54 & 9.88 \\
\midrule
\bf Many-to-Many & 8.56 & 7.97 & 7.59 & \\
 \bottomrule 
\end{tabular}
\caption{\textbf{Many-to-Many Performance by Domain.} We show spBLEU on three partitions of the  \flores{} devtest according to the originating domains. We compute the corpus spBLEU for each language in each domain, and then average across languages in that direction. We compute the performance into and out of English, Chinese, Spanish, Hindi, and Arabic, as well as average across all many-to-many directions. 
Overall, the News domain has slightly improved performance, but the domain effect is not strong.}
\label{tab:many2many_bleu_domain}
\end{table*}

\begin{table*}[ht]
\small
\centering
\begin{tabular}{lrrrrrrrrrrrr}
 & \bf \rotatebox{90}{Afro-Asiatic} & \bf \rotatebox{90}{Austronesian} & \bf \rotatebox{90}{Balto-Slavic} & \bf \rotatebox{90}{Bantu} & \bf \rotatebox{90}{Dravidian} & \bf \rotatebox{90}{Germanic} & \bf \rotatebox{90}{Indo-Aryan} & \bf \rotatebox{90}{{ \scriptsize  Nilotic+Other AC }} & \bf \rotatebox{90}{Romance} & \bf \rotatebox{90}{{ \scriptsize Sino-Tibetan+Kra-Dai }} & \bf \rotatebox{90}{Turkic} & \bf Avg\\
\cmidrule{2-12}
 \bf Num Languages: &  7 & 6 & 14 & 10 & 4 & 9 & 14 & 5 & 10 & 4 & 5 & \\
\midrule  
\bf Afro-Asiatic & \cellcolor{gray!25} 4.20 & 6.82 & 10.93 & 2.31 & 1.21 & \underline{11.95} & 3.43 & 0.93 & 11.70 & 3.66 & 2.73 & 5.44 \\
\bf Austronesian & 6.39 & \cellcolor{gray!25} \underline{11.50} & 13.78 & 3.48 & 2.08 & 15.53 & 4.69 & \bf 1.45 & 14.95 & 5.78 & 4.13 & 7.61 \\
\bf Balto-Slavic & 8.32 & 12.29 & \bf \cellcolor{gray!25} \underline{22.81} & 3.48 & 3.25 & 21.67 & 6.82 & 0.89 & 21.75 & 7.31 & 5.87 & 10.41 \\
\bf Bantu & 3.28 & 5.70 & 6.29 &\cellcolor{gray!25}  2.37 & 1.16 & \underline{7.40} & 2.16 & 1.37 & 7.16 & 2.77 & 1.95 & 3.78 \\
\bf Dravidian & 3.04 & 4.56 & 7.21 & 1.44 & \cellcolor{gray!25} 2.34 & \underline{7.66} & 3.62 & 0.39 & 7.31 & 2.73 & 2.04 & 3.85 \\
\bf Germanic & \bf 9.48 &\bf  14.25 & 22.56 & \bf 4.17 &\bf 3.26 &\bf  \cellcolor{gray!25} \underline{26.09} & \bf 6.89 & 1.40 &  23.53 & \bf 7.98 & \bf 6.24 & \bf 11.44 \\
\bf Indo-Aryan & 3.64 & 5.27 & 8.56 & 1.60 & 2.01 & \underline{8.81} & \cellcolor{gray!25} 3.70 & 0.42 & 8.66 & 3.26 & 2.36 & 4.39 \\
\bf { \scriptsize Nilotic+Other AC } & 1.60 & 3.10 & 2.76 & 1.45 & 0.43 & \underline{3.48} & 0.79 & \cellcolor{gray!25} 0.95 & 3.48 & 1.29 & 0.81 & 1.83 \\
\bf Romance & 8.25 & 12.74 & 20.70 & 3.22 & 2.41 & 22.43 & 6.04 & 1.15 & \bf \cellcolor{gray!25}\underline{24.44} & 6.96 & 5.46 & 10.35 \\
\bf { \scriptsize Sino-Tibetan+Kra-Dai } & 4.68 & 7.45 & 10.58 & 2.29 & 2.29 & 10.84 & 4.10 & 0.67 & \underline{11.05} & \cellcolor{gray!25}5.10 & 3.20 & 5.66 \\
\bf Turkic & 3.55 & 5.24 & \underline{9.35} & 1.61 & 1.24 & 8.81 & 2.96 & 0.58 & 9.14 & 3.13  & \cellcolor{gray!25}2.38 & 4.36 \\
\midrule
\bf Avg & 5.13 & 8.08 & 12.32 & 2.49 & 1.97 & 13.15 & 4.11 & 0.93 & 13.01 & 4.54 & 3.38 \\
 \bottomrule 
\end{tabular}
\caption{\textbf{Many-to-Many Performance on Family Groups}. We display the spBLEU on the devtest of \flores{} for the M2M-124 615M parameter model. We group languages into 11 families. Each cells represent the average performance for translating from all the languages in the source group (row) into the each language of the target group (column). We highlight in gray the cells that correspond to within group evaluation.  In bold we show the best performance per target group and underline the best performance per source group.}
\label{tab:many2many_bleu_subgrouping}
\end{table*}

\paragraph{English-Centric Translation.} Across the board, performance of translation \textit{into} English is strong, with only a few languages with spBLEU below 10. 
Performance \textit{out of} English is much worse.
We display this graphically in Figure~\ref{fig:res_vs_to_from_en_bleu}, where we show that performance into English (circle markers) is has higher spBLEU than performance out of English (triangle markers). 
Further, performance is overall heavily correlated with amount of training data, which we discuss in greater detail later.

\paragraph{Many-to-Many Translation.} Across non-English-Centric directions, performance requires improvement --- translation in and out of most African languages, for example, struggles to reach 5 spBLEU.
In contrast, translation into many European languages, even low-resource languages such as Occitan, have much better performance (over 10 spBLEU for many directions).
This result highlights the importance of both the amount of data and transfer learning from related languages.
For instance, translation to and from Occitan can naturally borrow from related high-resource languages like French, Italian and Spanish. However, the same cannot be said about most African languages, for which related languages are also low resource and difficult to translate.

\paragraph{Performance by Resource Level.}

A challenge of analyzing performance of various language families is that performance is often closely tied to the amount (and quality) of available training data. 
Certain language families have much less data. 
For example, almost every single African language is considered a low-resource translation direction. 
Thus, we next evaluate performance based on resource level. 
We classify languages into four bins based on resource level of bitext data with English: \textit{high}-resource languages, with more than 100M sentences of training data, \textit{mid}-resource with between 1M and 100M sentences, \textit{low}-resource with between 100K and 1M sentences, and finally \textit{very low}-resource with less than 100K sentences.

Our results are summarized in Table~\ref{tab:many2many_bleu_resources}. 
As hypothesized, performance increases with greater quantity of training data, in a clear pattern. spBLEU increases moving from left to right, as well as from top to bottom. 
Translation between mid and high resource languages produces spBLEU scores around 20, whereas translating between very low and low-resource languages yields a mere spBLEU score of less than 5. 
Even translation between high resource and low-resource languages is still quite low, indicating that lack of training data strongly limits performance of current MT systems. 

\begin{figure*}[t]
  \centering
  \includegraphics[width=1\linewidth]{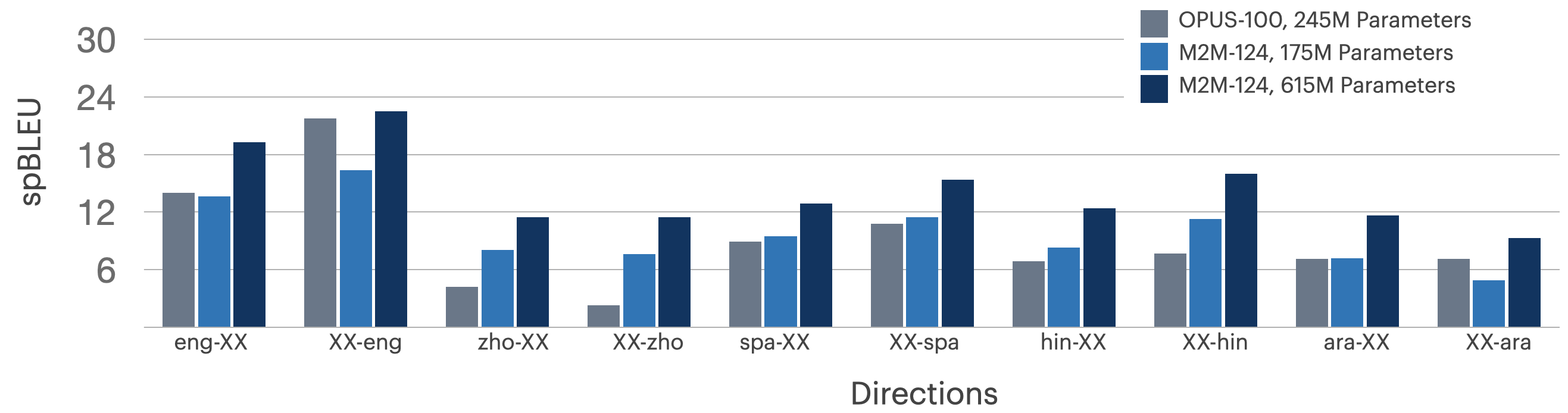}
  \caption{\textbf{Comparison between OPUS-100 and M2M-124} on several one-to-many and many-to-one translation tasks using five languages: English, Chinese, Spanish, Hindi, and Arabic. In each case, spBLEU is averaged over all languages in the set. Since the open-source OPUS-100 model covers only 80 languages of \flores{}, we restrict the evaluation to only these languages in order to make a fair comparison.}
   \label{fig:one2many_many2one}
\end{figure*}
 
\begin{figure*}[t]
  \centering
  \includegraphics[width=1\linewidth]{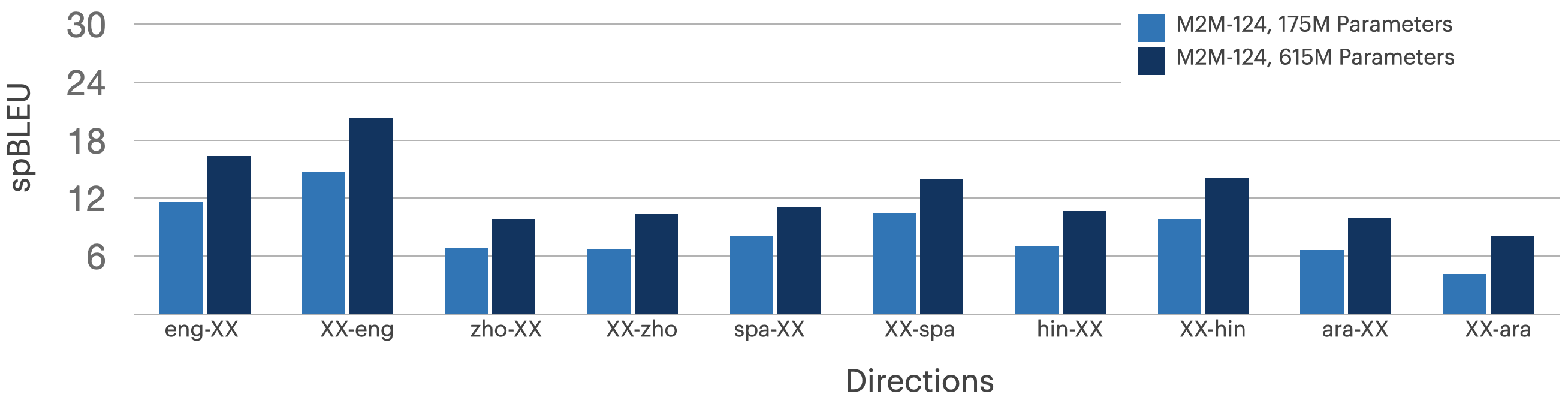}
  \caption{\textbf{Full results of M2M-124 Models} on several one-to-many and many-to-one translation tasks using five languages: English, Chinese, Spanish, Hindi, and Arabic. In each case, spBLEU is averaged over all languages in the set (all the remaining 100 languages of \flores{}).}
   \label{fig:one2many_many2one2}
\end{figure*}

\paragraph{Performance by Sentence Length.}

In the previous paragraphs we have found that translation quality is affected by the amount of training data and the properties of the language.
Next, we examine if translation quality is also affected a property of the sentences themselves. 
In particular, we calculate if the sentence length affects model performance, based on the hypothesis that longer sentences may be more complex and difficult to translate~\citep{sutskever2014sequence}. The results in Table~\ref{tab:many2many_bleu_sentence_length} show spBLEU on different subsets of the devtest, 
grouped by sentence length. The sentence length is determined by the number of tokens in the original English sentence. The bucket with short sentences collect all sentences with up to 15 tokens (in English), the medium bucket has sentences with a number of tokens in the range 16 to 25, and the last bucket has sentences with more than 25 tokens. The table shows that spBLEU is rather constant with respect to the sentence length, and in fact it slightly increases with the length of the sentence, contrary to our initial conjecture. 

\paragraph{Performance by Domain.}
We analyze if model performance is affected by domain, to check whether certain domains are more difficult to translate than others. 
\flores{} contains three domains: WikiNews, WikiJunior, and WikiVoyage.
We report results of translating in and out of five languages, namely English, Chinese, Spanish, Hindi, and Arabic, as well as the average across all of the $10{,}000$ possible directions.

The results in Table~\ref{tab:many2many_bleu_domain} demonstrate that the factor that affects the most translation quality is the language we translate in and out of (with Arabic being the most challenging and English having the highest scores) rather than the domain. WikiNews is the easiest domain with slightly higher spBLEU, and WikiVoyage is the hardest domain, with an average spBLEU score lower by one point compared to WikiNews. 
We hypothesize that news-related content is often written in a certain fairly consistent, journalistic style, which could ease the challenge of translation, while WikiVoyage might be a little harder because it has more named entities of local regions of the world which might be harder to translate correctly. However, overall, there are no large differences in performance between domains.

\paragraph{Performance by Language Family.}
We also group languages into eleven families based on general language families\footnote{Note: We define language subgroups for analysis purposes only. These are based on general language families, but are not completely aligned with an agreed upon linguistic taxonomy.} and report in Table~\ref{tab:all_languages} the average spBLEU for translating from and into each family.
Our results indicate that Bantu, Dravidian, Indo-Aryan, and Nilotic are the language families where M2M-124 struggles the most, attaining an average spBLEU below 5 points. 
In fact, even translation within the language family (see values in the diagonal) works very poorly. For these languages, translating to/from Germanic and Romance languages works better. In general, Germanic, Romance, and Balto-Slavic are the language families that yield the largest spBLEU scores (above 10 spBLEU points in average). For these latter languages translation within the language family works the best. In this case, M2M-124 obtains an spBLEU score above 20. 
Overall, translation between all languages in a many-to-many fashion requires improvement, as evidenced by the overall quite low average scores.

\subsubsection{Comparison of Various Systems.} 

We end by comparing various baseline systems, to understand the performance of some existing models on \flores{}.

\paragraph{Comparison to OPUS-100.} We evaluate OPUS-100~\cite{zhang-etal-2020-improving} with 254M parameters and the two versions of M2M-124~\cite{fan2020beyond} with 175 and 615M parameters. We calculate spBLEU in and out of five languages: English, Chinese, Spanish, Hindi, and Arabic.

Results are shown in Figure~\ref{fig:one2many_many2one}.
Note that OPUS-100 only covers 80 languages in \flores{}, so this figure is on the subset of 80 languages covered by all models, for comparability.  
Overall, we see a consistent trend across models and directions: the larger M2M-124 has the best performance, followed by the smaller M2M-124 and OPUS-100. For all systems we evaluated, translation to and from English works the best, while translation to and from Chinese and Arabic struggles the most.
In general, spBLEU scores are relatively low, suggesting ample room for improvement and need for further research in this area.

We next display results of M2M-124 175M parameters and 615M parameters on the full set of \flores{} languages. 
This is shown in Figure~\ref{fig:one2many_many2one2}. 
Comparing results with Figure~\ref{fig:one2many_many2one}, it is evident that the average performance in these language groupings has decreased, indicating that the additional languages in \flores{} are likely very difficult.
We see the same consistent trend that the larger M2M-124 model has stronger performance.

\paragraph{Comparison with Selected Masakhane Models.}

The comparison with OPUS-100 compares M2M-124 with another multilingual model. 
However, various researchers in the low-resource translation community have developed models for specific languages.
Many of these models are created by people who speak these languages. 
Further, focusing on specific directions of interest rather than multilingual models could produce specialized models with potentially higher quality. 

Masakhane is a participatory research effort that focuses on African NLP.
We end by comparing our M2M-124 model with several publicly available models from the \texttt{Masakhane-MT} repository. We evaluate models from English to the following languages: Yoruba\footnote{\url{https://github.com/masakhane-io/masakhane-mt/tree/master/benchmarks/en-yo/jw300-baseline-improve}},  Zulu\footnote{\url{https://github.com/masakhane-io/masakhane-mt/tree/master/benchmarks/en-zu/jw300-baseline}}, Swahili\footnote{\url{https://github.com/masakhane-io/masakhane-mt/tree/master/benchmarks/en-sw/fine-tuned-jw300-baseline}}, Shona\footnote{\url{https://github.com/masakhane-io/masakhane-mt/tree/master/benchmarks/en-sn/jw300-shona-baseline}}, Nyanja\footnote{\url{https://github.com/masakhane-io/masakhane-mt/tree/master/benchmarks/en-nya/jw-300-baseline}} and Luo\footnote{\url{https://github.com/masakhane-io/masakhane-mt/tree/master/benchmarks/en-luo/fine-tuned-jw300-baseline}}. The Masakhane models are trained on the JW300 dataset. 

Results are shown in Table~\ref{tab:masakhane_model_evaluation}. 
We observe that for two languages --- Zulu and Luo --- Masakhane's open sourced models have stronger performance on \flores{} than the M2M-124 model. The remaining languages we assess have either similar or worse performance than M2M-124. Overall, all languages besides Swahili require significant improvement. 
Note that in many African countries, a large number of local and regional languages are spoken. We hope that \flores{} can be used to develop non-English-centric models that directly translate between African languages.

\begin{table}[]
\small
\centering
\begin{tabular}{l|cc}
\toprule 
 & \bf Masakhane  & \bf M2M-124 \\
 \midrule
\bf English $\rightarrow$ Yoruba & 2.04 & 2.17 \\
\bf English $\rightarrow$ Zulu & 11.85 & 3.89 \\
\bf English $\rightarrow$ Swahili & 22.09 & 26.95 \\
\bf English $\rightarrow$ Shona & 8.19 & 11.74 \\
\bf English $\rightarrow$ Nyanja & 2.19 & 12.9 \\
\bf English $\rightarrow$ Luo & 5.33 & 3.37 \\
\bottomrule 
\end{tabular}
\caption{\textbf{spBLEU of various models open sourced by Masakhane-MT and M2M-124}. We evaluate models on translating from English to six different African languages. We compare against the M2M-124 615M parameter model.}
\label{tab:masakhane_model_evaluation}
\end{table}

\section{Conclusion}
\label{sec:conclusion}

The potential to develop translation systems for languages all over the world is hindered by lack of reliable, high quality evaluation.
Without the fundamental ability to measure translation quality, it is impossible to develop, iterate, and test various models and techniques. 
Particularly for low-resource languages where there is little available training data, new methods and algorithms  must be developed to improve translation of these languages. 
In this work, we create and open-source \flores{}, an evaluation benchmark covering 101 languages. 

\flores{} supports many-to-many evaluation, meaning any of 10,100 language directions can be evaluated. With rich metadata, it also supports multimodal translation via images, and document-level translation. Unlike many other multilingual datasets, \flores{} is fully translated by humans using a detailed process with numerous quality control checks, including human evaluation during dataset creation. 

Beyond translation, \flores{} can be used to evaluate tasks such as sentence and document classification, language identification, and multilingual domain adaptation. We hope that the release of this dataset and our baseline M2M models will be useful for the community. 

We hope to continue to expand the number of languages covered in \flores{} and make the test set available to various community efforts to improve translation systems in shared tasks such as those from the Workshop on Machine Translation. 

\section{Acknowledgments}

We'd like to thank Michael Auli and Sergey Edunov for enlightening discussions and advice. We'd like to thank Mona Diab and Denise Diaz for consulting on specific languages and providing invaluable guidance on translation quality. We'd like to thank Xian Li and Yuqing Tang for helping select the original sentences for translation as part of FLORES. Finally, we'd like to thank Brian Bui for helping with the organization of the data collection effort. We thank all of the translators and human evaluators, as well as the translation and quality assurance agencies we worked with, for helping create \flores{}. 
 
\clearpage

\clearpage

\bibliography{paper}
\bibliographystyle{acl_natbib}

\newpage
\clearpage
\appendix

\section{Translation Quality Guidelines}

Severities:
\begin{itemize}
\itemsep0em 
    \item \textbf{Critical Errors} are issues that render the content unfit for use. An error is only critical if it seriously distorts the meaning of the source, in such a way that it becomes completely incomprehensible or that the essence of the message is lost.
    \item \textbf{Major Errors} may confuse or mislead the user or hinder proper use of the product/service due to significant change in meaning or appear in a visible or important part of the content.
    \item \textbf{Minor Errors} don't lead to loss of meaning and wouldn't confuse or mislead the user but would be noticed, would decrease stylistic quality, fluency or clarity, or would make the content less appealing.
\end{itemize}

Error Categories:
\begin{enumerate}
\itemsep0em 
    \item \textbf{Grammar} --- Noncompliance with target language’s grammar rules. Grammar errors may be at the word or sentence level. Types of grammar errors may include:
    \begin{itemize}
    \itemsep0em 
        \item Incorrect person, number or case: the person, number or in the translation does not match the person, number or case of the source text.
        \item Incorrect tense: the tense used in the translation does not correspond to the tense used in the source.
        \item Incorrect subject/verb agreement: The subject and verb of a sentence must agree with one another in number whether they are singular or plural. If the subject of the sentence is singular, its verb must also be singular; and if the subject is plural, the verb must also be plural.
        \item Incorrect use of singular or plural: if a noun in the source text is plural, the corresponding noun and its qualifiers must be plural in the translation.
        \item Incorrect word order: word order of the translation is non-standard in the target language, or the translator has made a preferential change to the word order of the source.
    \end{itemize}
    \item \textbf{Punctuation} --- Punctuation is missing, non-standard in the target language or inconsistent with the source punctuation.
    \item \textbf{Spelling} --- Incorrect spelling in the target language. Types of spelling errors may include:
    \begin{itemize}
    \itemsep0em 
        \item Use of the wrong homophone for the context e.g. ‘bare with me’
        \item Typos
        \item Incorrect use of accents
    \end{itemize}
    \item \textbf{Capitalization} --- Noncompliance with target language rules e.g. not capitalising the start of a sentence or a proper noun.
    \item \textbf{Addition/Omission} --- An essential element from the source text is missing in the translation or unnecessary/superfluous elements are present in the translation but were not originally present in the source text.
    \item \textbf{Mistranslation} --- Source text has not been translated faithfully. Types of mistranslation errors may include:
    \begin{itemize}
    \itemsep0em 
        \item Incorrect interpretation of the source text
        \item Literal translations and mistranslations of phrasal verbs, rendering incorrect meaning in the target
        \item Lack of disambiguation of ambiguous terms
        \item Using a subpar word
    \end{itemize}
    \item \textbf{Unnatural Translation} ---  Translation does not sound natural to a native speaker of the target language. Source text is translated word for word, rendering the translation unnatural, or is grammatically correct but unnatural to a native speaker. 
    \item \textbf{Untranslated Text} --- Words are left untranslated from the source text. This is when there are words in the source language present in the translation which should have been translated into the target language.
    \item \textbf{Register} --- The style or register of the translation is inconsistent with the source and context.
\end{enumerate}

\section{Additional Results}

\paragraph{Details of Spectral Clusters}

The list of clusters formed by spectral clustering on spBLEU scores is shown in Table~\ref{tab:spectral_cluster_details}.

\begin{table*}[t]
\small
\centering
\begin{tabular}{l|l}
\toprule 
 \bf Cluster  & \bf Languages \\
\toprule 
\bf 1 & Assamese, Bengali, Hindi, Japanese, Korean, Malayalam, Marathi, Telugu, Urdu, Chinese \\ 
\midrule
\bf 2 & Asturian, Catalan, Spanish, Fula, French, Irish, Galician, Italian, Kabuverdianu, Lingala, \\
& Norwegian, Occitan, Portuguese, Turkish, Wolof \\
\midrule
\bf 3 & Cebuano, Indonesian, Icelandic, Javanese, Khmer, Lao, Malay, Nyanja, Pashto, Shona, \\ 
& Swahili, Thai, Tagalog, Vietnamese \\
\midrule
\bf 4 & Bulgarian, Bosnian, Czech, German, Luxembourgish, Macedonian, Romanian, Russian, Slovak, \\
& Slovenian, Swedish, Ukrainian  \\
\midrule
\bf 5 & Hausa, Igbo, Kamba, Luo, Maori, Northern Sotho, Oromo, Somali, Umbundu, Xhosa, Yoruba, Zulu \\
\midrule
\bf 6 & Azerbaijani, Belarusian, Greek, Estonian, Finnish, Hungarian, Kazakh, Kyrgyz, Lithuanian, Latvian, \\& Mongolian, Tajik, Uzbek  \\
\midrule
\bf 7 & Gujarati, Kannada, Nepali, Oriya, Punjabi, Sindhi, Tamil \\
\midrule
\bf 8 & Afrikaans, Amharic, Arabic, Welsh, Danish, English, Farsi, Armenian, Hebrew, Georgian, Kurdish, Maltese \\
\bottomrule 
\end{tabular}
\caption{\textbf{Language clusters after applying spectral clustering on the full spBLEU matrix}: Interestingly, the spectral clustering identifies several clusters that are reminiscent of world regions, where these languages are often spoken together.}
\label{tab:spectral_cluster_details}
\end{table*}

 \begin{figure}[t]
  \centering
  \includegraphics[width=1\linewidth]{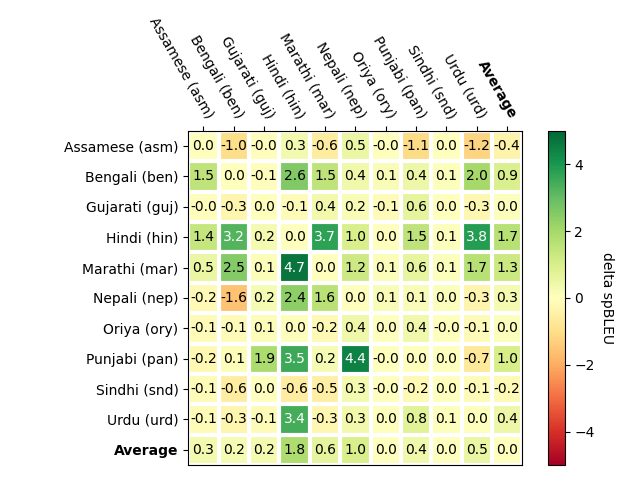}
  \caption{\textbf{Performance between Many-to-Many direction translation and English-Centric Pivoting.} We compare the difference in spBLEU (positive indicates direct translation has stronger performance) for 10 Indic languages. The results are computed using the M2M-124 615M model.
  }
   \label{fig:indic_langs_bridge_heatmap}
\end{figure}

\paragraph{Comparison of Many-to-Many with English-Centric Pivoting.}

We compare the need of evaluation in a truly many-to-many sense. 
Instead of creating multilingual models that can translate directly between any pair of languages, \textit{pivoting} through English is also possible.
Pivoting works by first translating from language X into English, then from English to language Y, instead of translating from X to Y.
\flores{} supports the evaluation and comparison of these strategies. 
Unlike previous work such as~\citet{fan2020beyond}, which was unable to evaluate all directions of their many-to-many model, \flores{} enables evaluation of all 101 x 101 pairs.

In Figure~\ref{fig:indic_langs_bridge_heatmap}, we compare direct translation with English-Centric Pivoting for 10 Indic languages: Assamese, Bengali, Gujarati, Hindi, Marathi, Nepali, Oriya, Punjabi, Sinhala, and Urdu. 
The spBLEU difference between direct translation and English pivoting is displayed in the heatmap. 
Overall, we see gains through 80\% of the directions by translating directly in a many-to-many fashion. 
Some directions have gains of more than 3 spBLEU, while a majority of the quality decrease from pivoting is less than 1 spBLEU.

\paragraph{tSNE of Model Embeddings.}

\begin{figure*}[ht]
\begin{center}
\includegraphics[width=0.9\linewidth]{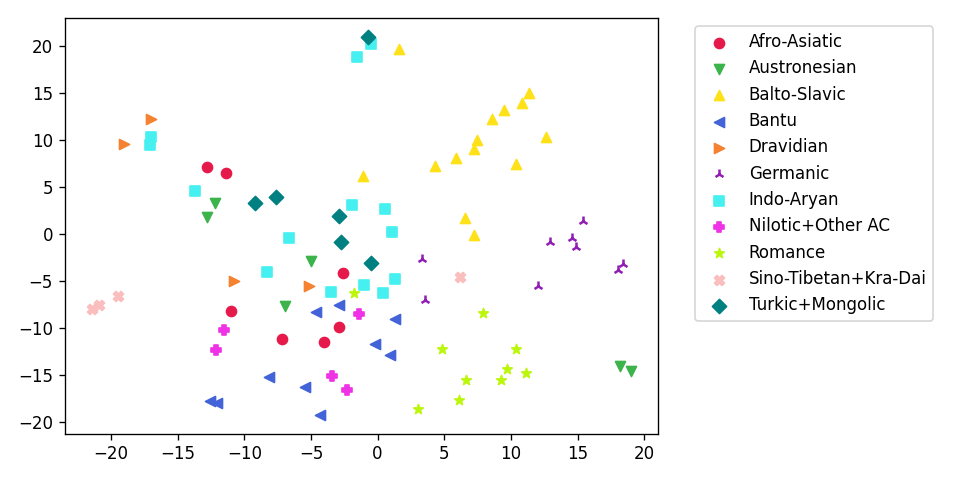}\qquad
\end{center}
\caption{\small \textbf{tSNE plot of Language Embeddings.} We embed the data of various languages with our model and examine by language subgrouping. Oftentimes, languages in the same subgrouping cluster together.}
\label{fig:lang_vector_tsne}
\end{figure*}

We examine the similarity of various languages by visualizing the tSNE of language embedding of the trained M2M-124 615M model. 
Unlike spectral clustering, this examination is a reflection of the model embeddings, rather than the spBLEU score.
Figure \ref{fig:lang_vector_tsne} shows that the languages belonging to the same language family are often grouped together, clustered next to each other.

\end{document}